\PassOptionsToPackage{table,xcdraw}{xcolor}

\documentclass[11pt]{article}

\usepackage[]{acl}

\usepackage{times}
\usepackage{latexsym}

\usepackage[T1]{fontenc}

\usepackage[utf8]{inputenc}

\usepackage{microtype}

\usepackage{inconsolata}

\usepackage{graphicx}
\usepackage{booktabs}
\usepackage{multirow}
\usepackage{makecell}
\usepackage{subfigure}
\usepackage[misc]{ifsym} 
\usepackage{balance}
\usepackage{listings}
\usepackage{color, xcolor}
\usepackage{algorithm2e}
\usepackage{tcolorbox}
\usepackage{amsmath}
\usepackage{cleveref}
\usepackage{url}
\usepackage{xurl}
\usepackage{hyperref}
\usepackage{titletoc}
\usepackage[toc,page,header]{appendix}
\usepackage{amsfonts}
\usepackage{amssymb}
\usepackage{listings}
\usepackage{xcolor}
\usepackage{subfigure}

\definecolor{lightcoral}{rgb}{0.94, 0.5, 0.5}
\definecolor{lightgreen}{rgb}{0.56, 0.93, 0.56}
\definecolor{harvestgold}{rgb}{0.85, 0.57, 0.0}
\definecolor{brightlavender}{rgb}{0.75, 0.58, 0.89}
\definecolor{capri}{rgb}{0.0, 0.75, 1.0}
\definecolor{carminepink}{rgb}{0.92, 0.3, 0.26}
\definecolor{celadon}{rgb}{0.67, 0.88, 0.69}
\definecolor{darkpastelgreen}{rgb}{0.01, 0.75, 0.24}
\definecolor{DeepSkyBlue4}{RGB}{0,104,139}

\crefformat{section}{\S#2#1#3} 
\crefformat{subsection}{\S#2#1#3}
\crefformat{subsubsection}{\S#2#1#3}

\title{MicroRemed: Benchmarking LLMs in Microservices Remediation}

\author{
  \textbf{Lingzhe Zhang\textsuperscript{1}$^{\dag}$},
  \textbf{Yunpeng Zhai\textsuperscript{2}$^{\dag}$},
  \textbf{Tong Jia\textsuperscript{1}$^{\ast}$},
  \textbf{Chiming Duan\textsuperscript{1}},
\\
  \textbf{Minghua He\textsuperscript{1,2}},
  \textbf{Leyi Pan\textsuperscript{3}},
  \textbf{Zhaoyang Liu\textsuperscript{2}},
  \textbf{Bolin Ding\textsuperscript{2}},
  \textbf{Ying Li\textsuperscript{1}\thanks{Corresponding Authors. $^{\dag}$Equal Contribution.}}
\\
  \textsuperscript{1}Peking University, China
  \textsuperscript{2}Alibaba Group, China
\\
  \textsuperscript{3}Tsinghua University, China
\\
 \texttt{\small zhang.lingzhe@stu.pku.edu.cn, zhaiyunpeng.zyp@alibaba-inc.com,} \\ 
 \texttt{\small jia.tong@pku.edu.cn, li.ying@pku.edu.cn}
}

\begin{document}
\maketitle

\begin{abstract}
Large Language Models (LLMs) integrated with agent-based reasoning frameworks have recently shown strong potential for autonomous decision-making and system-level operations. One promising yet underexplored direction is microservice remediation, where the goal is to automatically recover faulty microservice systems. Existing approaches, however, still rely on human-crafted prompts from Site Reliability Engineers (SREs), with LLMs merely converting textual instructions into executable code. To advance research in this area, we introduce MicroRemed, the first benchmark for evaluating LLMs in end-to-end microservice remediation, where models must directly generate executable Ansible playbooks from diagnosis reports to restore system functionality. We further propose ThinkRemed, a multi-agent framework that emulates the reflective and perceptive reasoning of SREs. Experimental results show that MicroRemed presents substantial challenges to current LLMs, while ThinkRemed improves end-to-end remediation performance through iterative reasoning and system reflection.\footnote{The benchmark is available at \url{https://github.com/LLM4AIOps/MicroRemed}.}
\end{abstract}

\section{Introduction}

Large Language Models (LLMs) have demonstrated remarkable advancements in recent years, exhibiting strong capabilities in understanding, reasoning, and problem-solving across a wide range of domains~\cite{guo2025deepseek, el2025competitive, zhang2025survey}. As research continues to expand beyond pure text generation, LLMs are increasingly being integrated with agent-based frameworks that enable autonomous decision-making and execution-oriented reasoning~\cite{zhang2025agentfm, singh2025agentic, pan2025omni}. These developments empower LLMs not only to generate natural language responses but also to interact with external environments, plan multi-step actions, and perform automatic operations in complex real-world settings.

In particular, software maintenance and operations have emerged as a promising frontier for LLM applications, where effective automation requires intensive interaction with software systems, such as querying runtime environments, interpreting diagnostic feedback, and executing repair actions~\cite{zhang2025survey, zhang2025thinkfl, liu2025ora}. Within this domain, microservice systems—as a dominant architectural paradigm in modern software—pose distinctive challenges for intelligent automation~\cite{duan2025logaction, he2025walk}. Their highly distributed and dynamic nature often leads to cascading failures that demand rapid and accurate auto-remediation~\cite{joel2024survey, sanwouo2025generative, trabelsi2024exploring}. Achieving reliable remediation in such environments requires both semantic understanding of system states and actionable reasoning over complex dependencies.

\begin{figure}[htbp]
	\centering
	\includegraphics[width=1\linewidth]{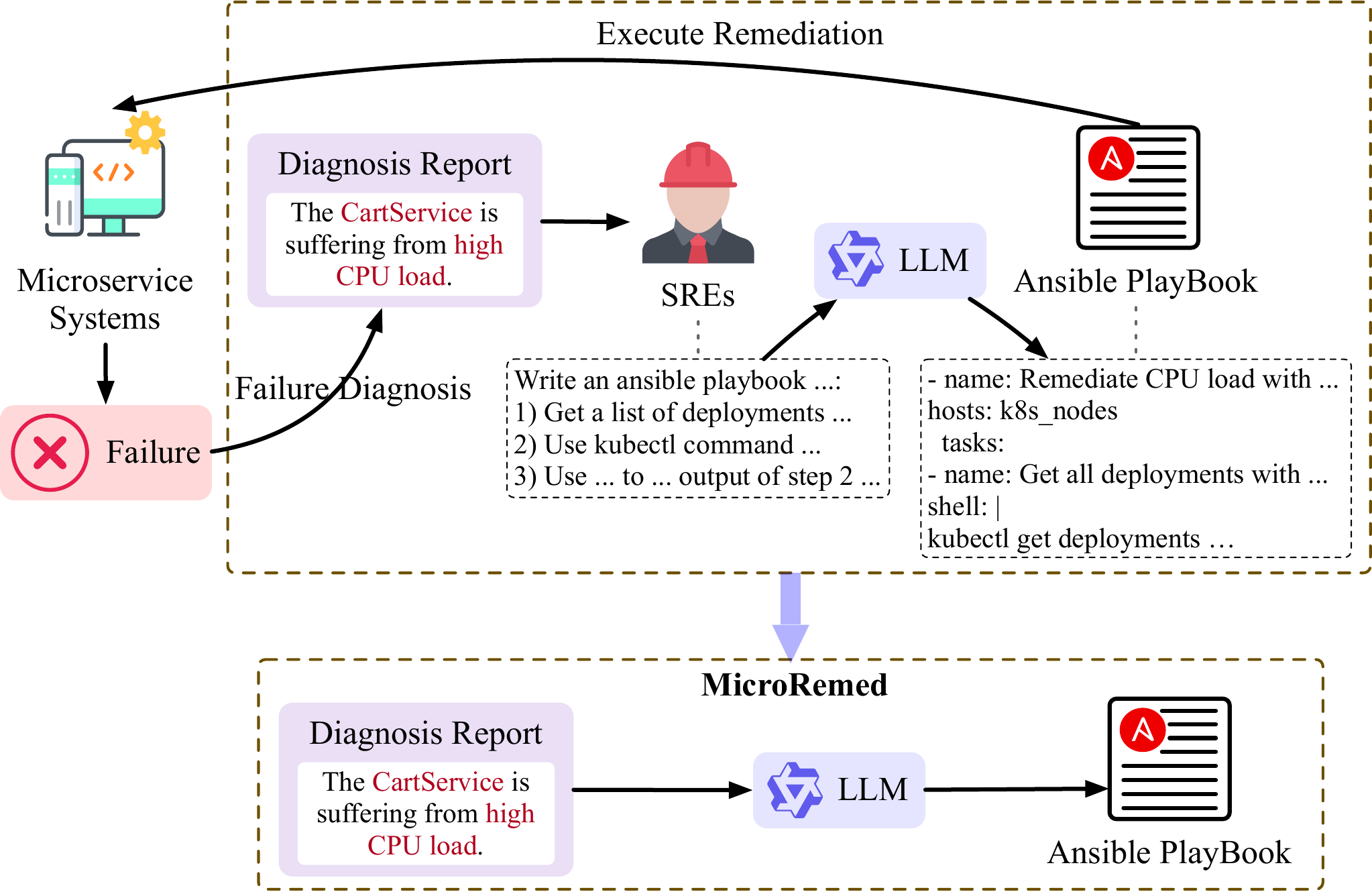}
	\caption{Previous microservice remediation workflow compared with the end-to-end microservice remediation pipeline proposed in MicroRemed.}
	\label{fig: intro-example}
\end{figure}

Previous work on microservice remediation mainly focuses on using LLMs to generate ansible playbooks that can be executed to repair faulty services. Representative approaches such as Wisdom-Ansible~\cite{pujar2023automated} and MAPE-Ansible~\cite{sarda2024leveraging} translate human-written instructions into executable remediation scripts. To support such studies, benchmarks like KubePlaybook~\cite{namrud2024kubeplaybook} and Andromeda~\cite{opdebeeck2021andromeda} provide curated collections of prompts and playbook templates for automation tasks. These datasets have facilitated the exploration of LLM-driven remediation capabilities.

However, as shown in Figure~\ref{fig: intro-example}, existing methods and benchmarks still face key limitations. Their remediation process typically depends on human-crafted prompts written by experienced SREs, where the LLM merely converts textual instructions into code. Such designs rely heavily on manual intervention, lack iterative feedback from the runtime environment, and fail to realize end-to-end automation from failure diagnosis to system recovery.

To fill this gap, we propose a new task, \textbf{End-to-End Microservice Remediation (E2E-MR)}, which aims to directly generate executable ansible playbooks from a given diagnosis report and autonomously recover the faulty system. This task establishes a closed-loop remediation pipeline, where LLMs translate diagnostic insights into concrete repair actions that can be automatically executed within the microservice environment. To evaluate this task, we introduce \textbf{MicroRemed}, a benchmark designed to assess LLMs’ capabilities in end-to-end microservice remediation. MicroRemed automatically deploys a real microservice system and continuously injects diverse failures. For each injected failure, it generates a corresponding diagnosis report based on the target component and failure type, which is then provided to the LLM under evaluation. The LLM produces an Ansible playbook, which is executed automatically, and the system subsequently verifies whether the injected failure has been successfully repaired. MicroRemed supports unlimited rounds of random failure injection and verification, allowing for extensive stress testing and iterative evaluation. Moreover, to facilitate fair and structured comparison, we categorize remediation targets into three difficulty levels—easy, medium, and hard—based on the complexity and interdependency of the underlying failure scenarios.

Additionally, we introduce two reference methodologies: \textbf{SoloGen} and \textbf{ThinkRemed}. SoloGen represents a pure one-shot generation approach. It takes as input all potentially relevant contextual information at once and prompts the language model to directly produce the final Ansible playbook in a single pass. In contrast, ThinkRemed is a multi-agent framework designed to emulate the SRE-like remediation process in microservice systems. It equips the model with a probe agent that enables dynamic information acquisition from the running system, guiding the model through iterative reasoning before finalizing the playbook. Moreover, ThinkRemed allows limited trial-and-reflection cycles, enabling the model to refine its plan based on playbook execution feedback and thereby mitigating the risk of incomplete or inaccurate decisions caused by missing information.

To validate the effectiveness and generality of MicroRemed, we integrate two widely used microservice systems—Train-Ticket~\cite{zhou2018fault} and Online-Boutique~\cite{google2025onlineboutique}—which are well recognized for simulating realistic production environments. In addition, we include a self-developed lightweight system, Simple-Micro, to provide controlled experiments and facilitate fine-grained analysis. We then evaluate nine representative LLMs, covering both closed-source and open-source ones, under the two proposed reference methodologies: SoloGen and ThinkRemed. The experimental results reveal that even the most capable LLMs still struggle to achieve satisfactory remediation performance on MicroRemed, underscoring the substantial challenges and realism of the benchmark.

The contributions of out work are as follows:
\begin{itemize}
	\item We construct a challenging benchmark, MicroRemed, which integrates seven automated failure injection and validation mechanisms and three realistic microservice systems, enabling the generation of 421 distinct fault–recovery pairs for evaluation.
	\item To address the challenge of end-to-end microservice remediation, we propose ThinkRemed, a collaborative multi-agent framework that performs dynamic probing, iterative reasoning, and limited trial-and-reflection cycles to generate effective remediation actions.
	\item Extensive experiments across nine large language models demonstrate that MicroRemed poses substantial challenges to existing LLMs. Moreover, ThinkRemed’s ability to perceive and reflect on system states can enhance the performance of end-to-end microservice remediation.
\end{itemize}

\section{Related Work}

\subsection{LLM-based Software Maintenance}

Automatic software maintenance and operations have long been an active area of research~\cite{}. From a system lifecycle perspective, this process is typically divided into three progressive stages: anomaly detection, failure diagnosis, and software remediation. Anomaly detection focuses on identifying whether abnormal behaviors have occurred within the system. Once an anomaly is detected, failure diagnosis aims to localize the root cause and characterize the nature of the failure. Finally, software remediation builds upon diagnostic results to execute appropriate recovery actions and restore the system to a healthy state.

Recently, LLMs have been increasingly integrated into these stages to enhance automation and reasoning capabilities. In LLM-based anomaly detection, existing research has explored both fine-tuning foundation models on structured telemetry data such as metrics and logs~\cite{ning2025ts, das2024decoder, le2024prelog} and developing prompt-driven methods that directly leverage pre-trained LLMs for unsupervised anomaly identification~\cite{caotempo, duan2025eagerlog, liu2024lstprompt}.

In LLM-based failure diagnosis, studies can be broadly categorized into two lines: (i) failure localization, which employs multi-agent frameworks to identify the component where a failure occurs~\cite{pei2025flow, li2025coca, zhang2025thinkfl, zhang2025adaptive}, and (ii) failure classification, which leverages retrieval-augmented reasoning~\cite{zhang2025scalalog, zhang2024lm} or similar paradigms to infer the underlying cause of failures.

In contrast, LLM-based software remediation remains in its early stage.
Most existing work focuses on mitigation solution generation~\cite{ahmed2023recommending, jiang2024xpert}, which suggests potential manual recovery actions to SREs. A few recent studies attempt to generate executable Ansible playbooks based on SRE-provided prompts~\cite{pujar2023automated, sarda2024leveraging, namrud2024kubeplaybook}, bridging natural language reasoning with system repair scripts. However, these approaches still rely heavily on human intervention and lack true end-to-end automation.

\subsection{Software Remediation Benchmark}

Software remediation, involving complex processes of generation, analysis, and execution, has only become feasible in the LLM era. A representative benchmark in this direction is SWE-Bench~\cite{jimenezswe, yangswe}, which contains 2,294 software engineering tasks derived from GitHub issues and pull requests across 12 Python repositories. However, SWE-Bench mainly targets bug fixing in software development, rather than runtime system remediation, which requires dynamic diagnosis and execution in operational environments. In the field of intelligent operations, AIOpsLab~\cite{ma2025aiopslab} provides a framework for building and evaluating autonomous AIOps agents, yet it lacks standardized and reproducible evaluation mechanisms for verifying remediation effectiveness. In the microservice domain, KubePlaybook~\cite{namrud2024kubeplaybook} and Andromeda~\cite{opdebeeck2021andromeda} offer curated prompts and playbook templates for automation, but their remediation processes depend on human-crafted prompts, where the LLM merely translates instructions into executable scripts. This human-in-the-loop dependency limits reproducibility and scalability for end-to-end automated remediation evaluation.

\begin{figure*}[tbp]
	\centering
	\includegraphics[width=1\linewidth]{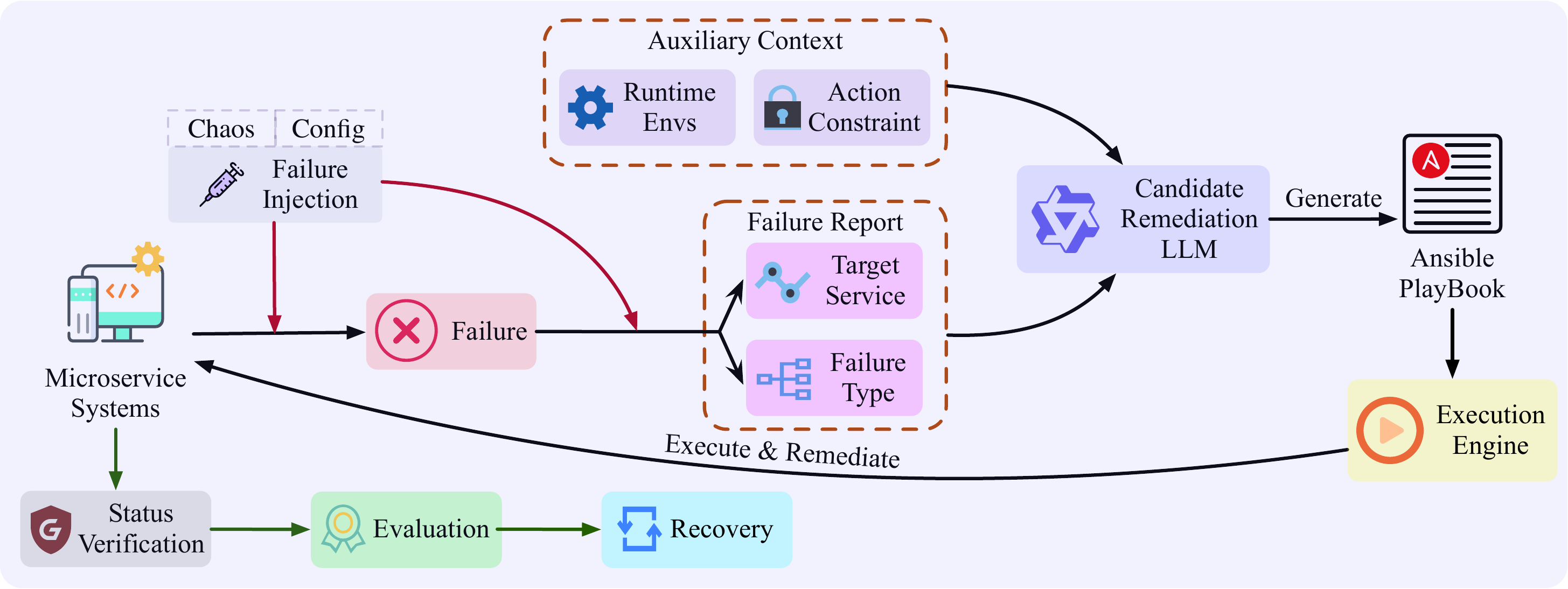}
	\caption{MicroRemed Benchmark Pipeline: the benchmark launches a real microservice; Failure Injection injects faults and produces a Failure Report; the Failure Report together with Auxiliary Context is provided to the Candidate Remediation LLM which generates an Ansible Playbook; the Execution Engine executes the playbook; Status Verification checks remediation success; Evaluation and Recovery restores the system for the next run.}
	\label{fig: benchmark}
\end{figure*}

\section{Benchmark Construction}

We present the construction of the MicroRemed benchmark in this section. We begin with an overview of the task definition and the underlying design principles (\cref{sec: design-principles}), followed by the architecture of the MicroRemed benchmark (\cref{sec: architecture}) and the evaluation protocol (\cref{sec: evaluation-protocol}). Finally, we describe the overall composition of MicroRemed (\cref{sec: benchmark-composition}).

\subsection{Design Principles}
\label{sec: design-principles}

Existing microservice remediation approaches typically depend on human-crafted prompts designed by experienced SREs, where LLMs merely translate natural language instructions into executable scripts such as Ansible playbooks. This paradigm lacks autonomy and generalization, as it relies heavily on explicit human reasoning rather than the model’s understanding of the system state.

To address this limitation, we introduce the task of \textbf{End-to-End Microservice Remediation (E2E-MR)}, which aims to evaluate an LLM’s ability to autonomously generate executable remediation plans given only structured diagnostic information. Unlike conventional prompt-based generation, E2E-MR emphasizes a direct remediation process that transforms diagnostic reports into actionable repair operations.

\begin{equation}
	\left\{
	\begin{aligned}
		f_\theta &: (\mathcal{S}_{target}, \mathcal{T}_{fail}, \mathcal{C}_{aux}) \to p^*,\\
		p^* &= \arg\max_{p \in \mathcal{P}} 
		\mathcal{U}\big(\mathcal{E}(p, \mathcal{S}_{fail}) = \mathcal{S}_{normal}\big)
	\end{aligned}
	\right.
	\label{eq: e2e-mr}
\end{equation}

Formally, the E2E-MR task can be formulated as Equation~\ref{eq: e2e-mr}, where $f_\theta$ is the candidate remediation LLM parameterized by $\theta$, $\mathcal{S}_{target}$ denotes the failed microservice, $\mathcal{T}_{fail}$ the failure type, and $\mathcal{C}_{aux}$ auxiliary contextual information. $\mathcal{P}$ is the space of executable playbooks, $\mathcal{E}$ represents the execution environment, and $\mathcal{U}(\cdot)$ measures the utility of successful recovery. The goal is to generate an optimal playbook $p^*$ that maximizes the likelihood of recovering the system state $\mathcal{S}_{fail}$ to $\mathcal{S}_{normal}$.

Therefore, to design a benchmark for the E2E-MR task, we adhere to the following design principles:

\begin{itemize}
	\item \textbf{Dynamic Execution Benchmark.}
	Unlike most LLM benchmarks that collect static data to form fixed datasets, the proposed benchmark is designed as a live and interactive execution environment. It actively launches real microservice systems, injects controlled failures, and interacts dynamically with running services. This design enables the benchmark to capture real-time behaviors, system dynamics, and contextual dependencies that static datasets cannot represent.
	
	\item \textbf{Execution-based Evaluation.}  
	Evaluation is not determined by linguistic or structural similarity of generated outputs, but by execution outcomes. Each generated playbook is executed within the microservice environment, and the benchmark verifies success by assessing whether the system has been fully recovered to its normal operational state.
	
	\item \textbf{Comprehensive Scalability.}  
	Built on these foundations, the benchmark is designed to be method-scalable, LLM-scalable, failure-scalable, and system-scalable. It supports diverse LLM-based remediation methods, allows plug-and-play replacement of remediation models, accommodates various failure scenarios, and can be easily extended to new microservice systems with minimal configuration effort.
\end{itemize}

\subsection{Architecture}
\label{sec: architecture}

Based on the above design principles, we develop MicroRemed. The overall architecture of MicroRemed is illustrated in Figure~\ref{fig: benchmark}. MicroRemed actively launches real microservice systems and performs Failure Injection to introduce controlled faults. According to the injected target service and failure type, it generates a Failure Report, which—together with a set of Auxiliary Contexts—is provided to the Candidate Remediation LLM to produce an executable Ansible Playbook. The playbook is then executed by an Execution Engine to carry out automated remediation. After execution, a Status Cerification module checks whether the issue has been successfully resolved.
Finally, the Evaluation and Recovery stage assesses the remediation outcome and restores the microservice system to its original state, thereby enabling reproducible and iterative experimentation.

The Failure Injection module introduces faults into the system through two complementary approaches: chaos injection and configuration injection. For resource-related or runtime failures (e.g., CPU stress, memory pressure, or network latency), MicroRemed adopts chaos injection, which dynamically perturbs the runtime environment using Chaos Mesh~\cite{mesh2025powerful} to emulate realistic fault conditions. For configuration-related failures (e.g., incorrect environment variables or service dependency misconfigurations), the system applies configuration injection, which directly modifies specific configuration files or environment settings to trigger controlled failures.

The Status Verification module resembles traditional anomaly detection in purpose but differs fundamentally in mechanism. While anomaly detection infers abnormality from large volumes of complex runtime data, status verification performs targeted validation of whether a specific injected failure has been fully remediated.
For example, if a CPU-stress failure was injected into service A, status verification will exclusively inspect the CPU metrics of service A to confirm recovery. This targeted design ensures 100\% verification accuracy, a level of precision unattainable by general anomaly detection approaches.

\subsection{Evaluation Protocol}
\label{sec: evaluation-protocol}

MicroRemed supports comprehensive evaluation from multiple perspectives, including performance, efficiency, and resource utilization. Specifically, we adopt the following metrics to quantify the effectiveness of candidate remediation LLMs:

\textbf{Remediation Accuracy (\textit{RA})} — measures the proportion of failures that are successfully repaired, reflecting the overall performance of the model.

\textbf{Average Remediation Latency (\textit{ARL})} — evaluates the temporal efficiency of each successful remediation cycle, encompassing both reasoning and execution delays.

\textbf{Average Token Consumption (\textit{ATC})} — quantifies the language-model cost efficiency, representing the average number of tokens consumed to achieve a successful remediation.

\subsection{Benchmark Composition}
\label{sec: benchmark-composition}

Although MicroRemed is designed with comprehensive scalability and supports extensible failure types and microservice systems, in our benchmark we include seven representative types of failures and three real-world microservice systems.

\begin{table}[htb]
	\setlength{\tabcolsep}{4.5pt}
	\centering
	\caption{Benchmark statistics on failure types}
	\label{tab: failure-types}
	\begin{tabular}{ccc}
		\toprule
		No. & Category & Failure Types \\
		\midrule
		1 & \multirow{3}*{Resource-Level} & CPU Saturation \\
		2 & ~ & Memory Saturation \\
		3 & ~ & IO Saturation \\
		\midrule
		4 & \multirow{2}*{Network-Level} & Network Loss \\
		5 & ~ & Network Delay \\
		\midrule
		6 & \multirow{2}*{Application-Level} & Pod Failure \\
		7 & ~ & Configuration Error \\
		\bottomrule
	\end{tabular}
\end{table}

\textbf{Failure Types.} As shown in Table~\ref{tab: failure-types}, MicroRemed includes seven representative failures across three categories: resource-level (CPU, memory, I/O saturation), network-level (network loss, network delay), and application-level (pod failure, configuration error).

\textbf{Microservice Systems.} MicroRemed integrates three microservice systems. Among them, two widely used benchmarks—Train-Ticket~\cite{zhou2018fault} and Online-Boutique~\cite{google2025onlineboutique}—are well recognized for emulating realistic production environments. In addition, we include a self-developed lightweight system, Simple-Micro, designed to enable controlled experiments and facilitate fine-grained analysis.

\textbf{Difficulty Levels.} Although MicroRemed supports arbitrary combinations of injected failures, we define three standardized difficulty levels—easy (23 cases), medium (49 cases), and hard (80 cases)—to enable fair and structured comparison across remediation methods. Each level corresponds to a curated set of failure combinations that vary in fault diversity, dependency complexity, and recovery difficulty.

\begin{figure*}[tbp]
	\centering
	\includegraphics[width=1\linewidth]{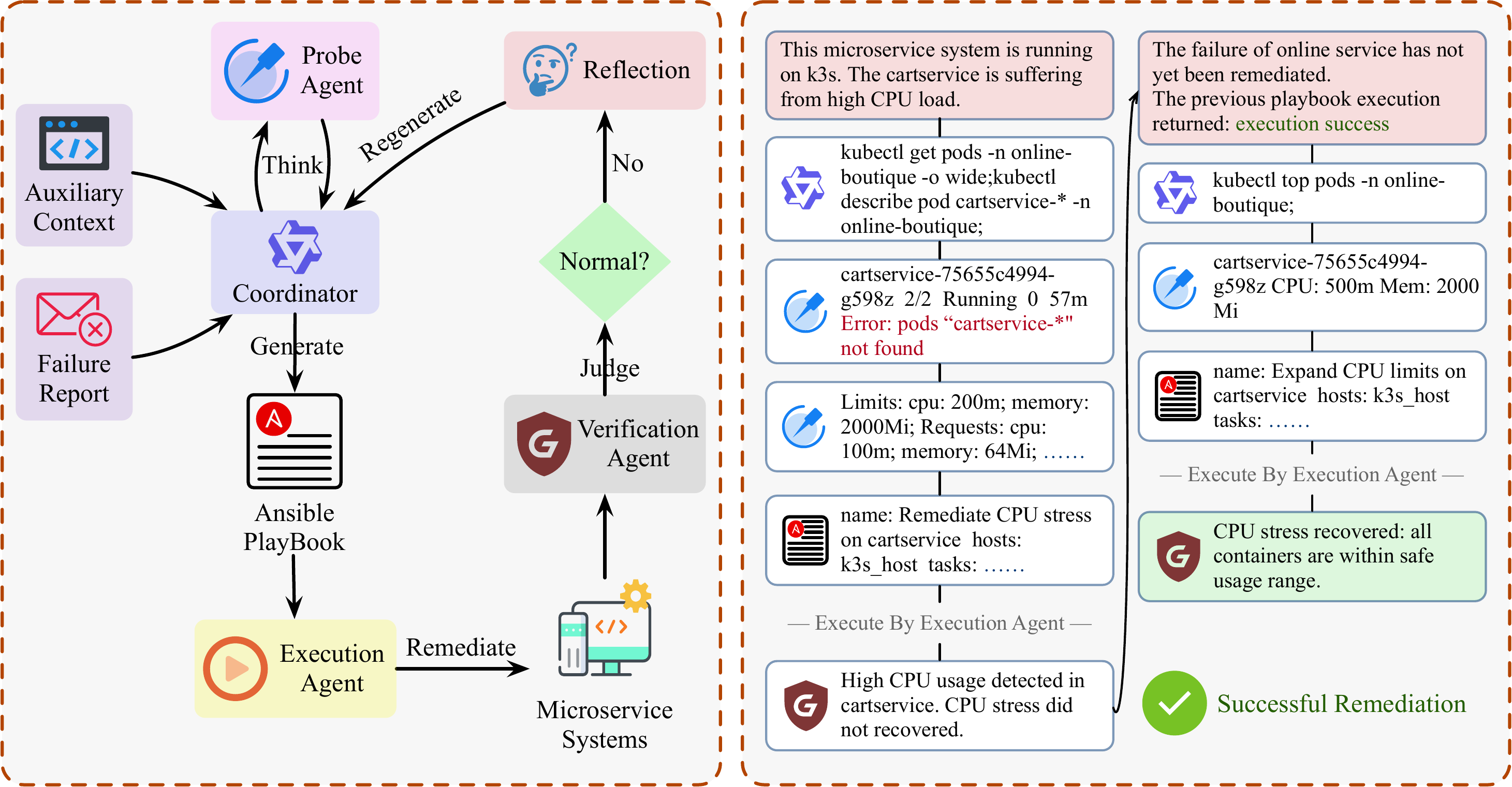}
	\caption{The overall framework of \textbf{ThinkRemed}}
	\label{fig: ThinkRemed}
\end{figure*}

\section{Reference Methodology}

To facilitate fair evaluation and comparison across different LLMs, we introduce two reference methodologies: \textbf{SoloGen} (\cref{sec: solo-gen}) and \textbf{ThinkRemed} (\cref{sec: think-remed}).

\subsection{SoloGen} \label{sec: solo-gen}

SoloGen represents a straightforward one-shot generation baseline. It replaces the candidate remediation LLM in the MicroRemed pipeline with a pre-trained large language model that receives all relevant contextual information in a single prompt and directly outputs the final Ansible playbook. This approach eliminates intermediate reasoning or iterative refinement, serving as a minimal yet effective baseline for debugging and evaluating the benchmark setup.

\subsection{ThinkRemed} \label{sec: think-remed}

While SoloGen performs direct generation without adaptive reasoning, it often struggles with complex multi-service dependencies and incomplete contextual information. To address these limitations, we propose ThinkRemed, a multi-agent framework designed to emulate the SRE-like remediation process in microservice systems.

As illustrated in Figure~\ref{fig: ThinkRemed}, ThinkRemed comprises four cooperative agents—\textit{Coordinator}, \textit{Probe}, \textit{Execution}, and \textit{Verification}—that operate within an iterative reasoning–action–reflection loop. The \textit{Coordinator} first receives the auxiliary context $\mathcal{C}_0$ and failure report $\mathcal{R}0$, and adaptively determines whether to invoke the \textit{Probe Agent} to collect additional runtime information from the system. It then synthesizes a candidate Ansible playbook $p_t$, which is executed by the \textit{Execution Agent} to remediate the faulty microservice system. The \textit{Verification Agent} subsequently assesses the remediation result, producing a binary outcome $v_t \in {0,1}$ indicating success or failure. If remediation fails, the system enters a reflection phase, and control returns to the \textit{Coordinator} for iterative refinement based on feedback. To ensure timely remediation and accommodate LLM context limitations, the iteration loop is bounded by a maximum trial budget $T{\max}$.

\begin{equation}
	\left\{
	\begin{aligned}
		p_t &= f_\theta(\mathcal{R}_t, \mathcal{C}_t, \mathcal{I}_t), \\
		s_{t+1} &= \mathcal{E}(p_t, s_t), \\
		v_t &= \mathcal{V}(s_{t+1}), \\
		(\mathcal{R}_{t+1}, \mathcal{C}_{t+1}) &= \mathcal{U}(\mathcal{R}_t, \mathcal{C}_t, s_{t+1}) \\
		&\qquad \text{if } v_t = 0 \text{ and } t < T_{\max}.
	\end{aligned}
	\right.
	\label{eq: think-remed}
\end{equation}

Formally, the iterative process of ThinkRemed can be represented as Equation~\ref{eq: think-remed}, where $f_\theta$ denotes the Coordinator’s reasoning policy, $\mathcal{E}$ the execution operator, $\mathcal{V}$ the verification predicate.

\begin{table*}[t]
	\setlength{\tabcolsep}{1.3pt}
	\centering
	\begin{tabular}{c|c|ccc|ccc|ccc|c}
		\toprule
		\multirow{2}{*}{\textbf{Backbone}} & \multirow{2}{*}{\textbf{Method}} & \multicolumn{3}{c|}{Train-Ticket} & \multicolumn{3}{c|}{Online-Boutique} & \multicolumn{3}{c|}{Simple-Micro} & \multirow{2}{*}{\textbf{Overall}} \\
		\\[-2.5ex]
		\cline{3-11}
		\\[-2ex]
		~ & ~ & \textit{Easy} & \textit{Medium} & \textit{Hard} & \textit{Easy} & \textit{Medium} & \textit{Hard} & \textit{Easy} & \textit{Medium} & \textit{Hard} & ~ \\
		\midrule
		\multicolumn{12}{c}{\cellcolor{brightlavender!15} \textbf{\textit{Closed-Sourced LLMs}}} \\
		\midrule
		\multirow{2}{*}{Qwen3-Plus} & SoloGen & 39.13 & 33.33 & 20.51 & 30.43 & 35.42 & 20.51 & 30.43 & 36.73 & 26.15 & 30.29 \\
		~ & ThinkRemed & 47.83 & 30.61 & 31.17 & 43.48 & 43.75 & 31.58 & 47.83 & 36.17 & 38.03 & \textbf{38.94} \\
		\midrule
		\multirow{2}{*}{Qwen3-Max}
		~ & SoloGen & 34.78 & 24.49 & 21.25 & 30.43 & 37.50 & 22.08 & 21.74 & 29.79 & 18.99 & 26.78 \\
        ~ & ThinkRemed & 47.83 & 28.57 & 30.77 & 39.13 & 37.50 & 25.32 & 30.43 & 22.92 & 17.91 & 31.15 \\
		\midrule
		\multirow{2}{*}{Qwen3-Flash}
		~ & SoloGen & 8.70 & 8.16 & 5.13 & 17.39 & 14.58 & 7.59 & 8.70 & 6.12 & 3.95 & 8.92 \\
		~ & ThinkRemed & 21.74 & 16.33 & 13.16 & 34.78 & 30.61 & 21.33 & 22.72 & 14.58 & 8.86 & 20.46 \\
		\midrule
		\multicolumn{12}{c}{\cellcolor{capri!15} \textbf{\textit{Open-Sourced LLMs}}} \\
		\midrule
		\multirow{2}{*}{QwQ-32B}
		~ & SoloGen & 4.35 & 6.38 & 6.41 & 13.04 & 12.50 & 10.13 & 8.70 & 10.20 & 8.70 & 8.93 \\
		~ & ThinkRemed & 17.39 & 10.20 & 7.89 & 26.09 & 22.45 & 15.58 & 17.39 & 8.33 & 6.76 & 14.68 \\			
		\midrule	
		\multirow{2}{*}{Qwen3-Next}
		~ & SoloGen & 8.70 & 10.20 & 6.49 & 19.05 & 15.56 & 12.00 & 21.74 & 12.50 & 1.47 & 11.97 \\
		~ & ThinkRemed & 13.04 & 6.12 & 5.06 & 17.39 & 17.02 & 17.72 & 21.74 & 28.57 & 19.35 & 16.22 \\
		\midrule			
		\multirow{2}{*}{Qwen3-235B}
		~ & SoloGen & 21.74 & 32.65 & 14.49 & 39.13 & 33.33 & 24.05 & 26.09 & 18.75 & 19.70 & 25.55 \\
		~ & ThinkRemed & 39.13 & 34.69 & 33.78 & 39.13 & 34.69 & 33.33 & 34.78 & 36.73 & 32.39 & \underline{35.41} \\					
		\midrule				
		\multirow{2}{*}{DeepSeek-V3.2}
		~ & SoloGen & 17.39 & 8.16 & 9.09 & 30.43 & 36.73 & 23.38 & 21.74 & 10.42 & 16.44 & 21.72 \\
		~ & ThinkRemed & 8.70 & 16.33 & 11.54 & 31.82 & 21.28 & 22.78 & 21.74 & 29.17 & 20.00 & 20.37 \\
		\midrule	
		\multirow{2}{*}{Kimi-K2}
		~ & SoloGen & 21.74 & 10.87 & 9.72 & 27.27 & 21.74 & 17.72 & 34.78 & 35.42 & 33.82 & 23.68 \\
		~ & ThinkRemed & 21.74 & 20.00 & 29.49 & 22.73 & 26.53 & 30.38 & 47.83 & 44.89 & 43.75 & 31.93 \\
		\midrule
		\multirow{2}{*}{GLM-4.5}
		~ & SoloGen & 13.04 & 23.40 & 12.99 & 39.13 & 24.49 & 15.79 & 34.78 & 16.33 & 16.25 & 21.80 \\
		~ & ThinkRemed & 21.74 & 20.41 & 27.63 & 43.48 & 43.75 & 39.47 & 43.48 & 36.73 & 30.38 & 34.12 \\
        \midrule
        \multicolumn{2}{c|}{\textbf{Overall}} & 22.71 & 18.94 & 16.48 & 30.24 & 28.81 & 20.41 & 27.59 & 24.13 & 21.37 & - \\
		\bottomrule
	\end{tabular}
    \caption{Remediation accuracy across closed-source and open-source LLM backbones}
	\label{tab: accuracy}
\end{table*}

\section{Experiment}

\subsection{Experimental Setting}

\textbf{Backbones.} To comprehensively evaluate the end-to-end microservice remediation capability of current LLMs, we examine a total of nine representative models, encompassing both closed-source and open-source variants.

\textit{Closed-Source LLMs:} Qwen3-Plus, Qwen3-Max, and Qwen3-Flash~\cite{yang2025qwen3}.

\textit{Open-Source LLMs:} QwQ-32B, Qwen3-Next-80B-A3V, Qwen3-235B-A22B, DeepSeek-V3.2-Exp~\cite{liu2024deepseek}, Kimi-K2~\cite{team2025kimi}, and GLM-4.5~\cite{zeng2025glm}.

\noindent \textbf{Implementation Details.} Considering that some models require excessively long reasoning time, we set the maximum thinking time to 5 minutes. If no result is returned within this limit, the attempt is regarded as a failure. Moreover, given the context length limitations of current models, we set the maximum retry number $T_{max}$ of ThinkRemed to 1, unless otherwise specified.

\subsection{Main Results}

The main results across nine LLMs are presented in Table~\ref{tab: accuracy}. In this table, the reported remediation accuracy only includes successfully injected failures.

As observed, Qwen3-Plus achieves the best performance at the model level, followed by Qwen3-235B. At the microservice level, Train-Ticket proves to be the most challenging environment, followed by Simple-Micro. Furthermore, ThinkRemed consistently outperforms SoloGen, with an average improvement of approximately 7.07\%. However, it is worth noting that even under the easiest level of the benchmark, ThinkRemed fails to reach 50\% accuracy, highlighting the overall difficulty and rigor of the MicroRemed benchmark.

\subsection{Latency–Accuracy Trade-Off}

\begin{figure}[htbp]
	\centering
	\includegraphics[width=1\linewidth]{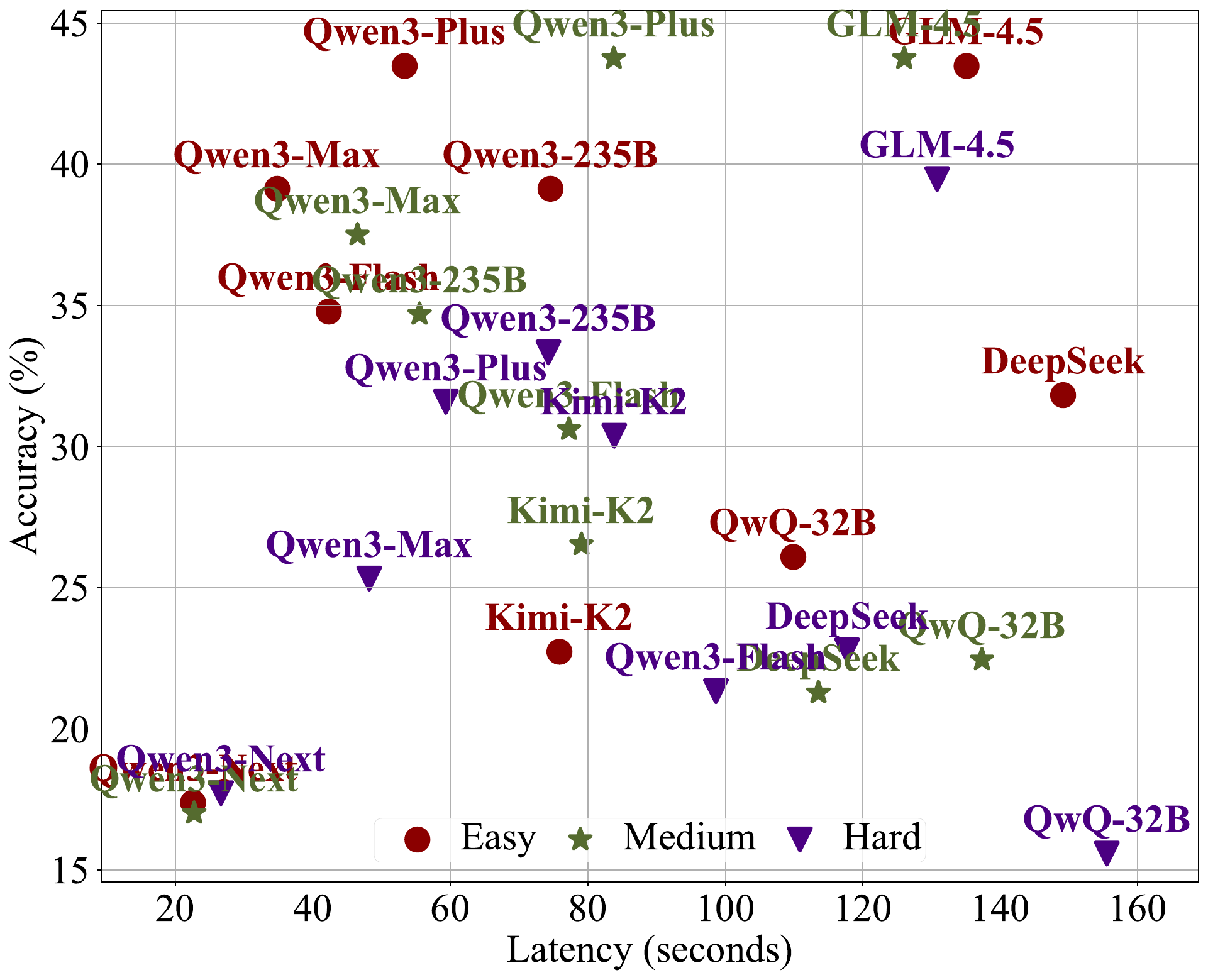}
	\caption{Latency–accuracy trade-off of various large language models across three difficulty levels (Easy, Medium, Hard) on the Online-Boutique microservice}
	\label{fig: latency-accuracy-capacity}
\end{figure}

The latency-accuracy trade-off of various large language models is shown in Figure~\ref{fig: latency-accuracy-capacity}. We reported three difficulty levels (Easy, Medium, Hard) on the Online-Boutique microservice. Each point represents a model, with its x-coordinate indicating average inference latency (seconds, lower is better) and y-coordinate showing accuracy (\%).

Models are grouped by task difficulty, represented with distinct markers and colors. Notably, Qwen3-Plus achieves the highest accuracy while maintaining relatively low latency. In contrast, Qwen3-Next and Qwen3-Flash, which exhibit even lower latency than Qwen3-Plus, show significantly reduced accuracy. Qwen3-Max delivers slightly lower accuracy than Qwen3-Plus but with marginally reduced latency. Models are grouped by task difficulty, represented with distinct markers and colors. It is also worth noting that QwQ-32B, despite its smaller model size, demonstrates high latency due to its enforced reasoning process; however, this forced reasoning does not lead to improved accuracy in this benchmark.

\subsection{Failure-Type-Wise Comparison}

\begin{figure}[htbp]
	\centering
	\includegraphics[width=1\linewidth]{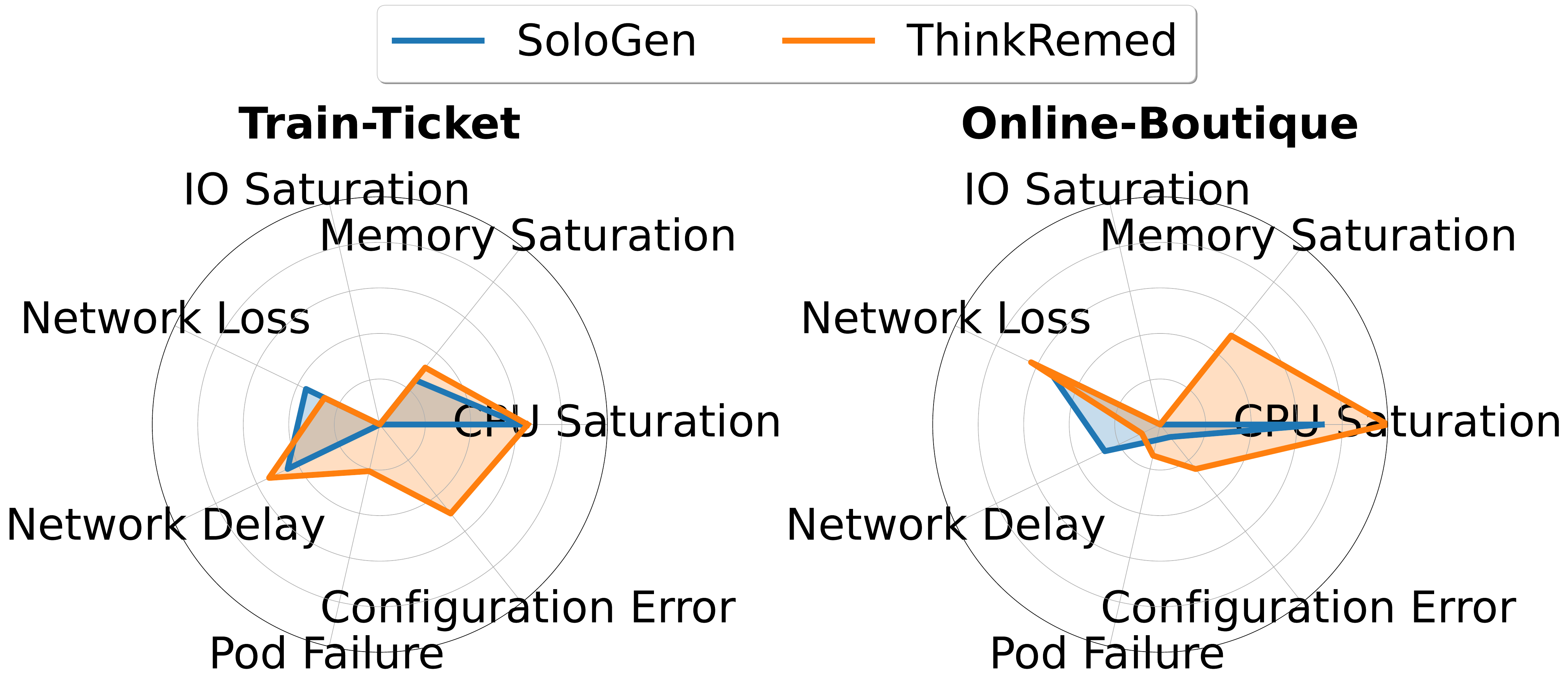}
	\caption{Class-wise performance comparison across failure types on the Train-Ticket and Online-Boutique}
	\label{fig: class-wise-comparison}
\end{figure}

We further analyze the remediation accuracy of SoloGen and ThinkRemed across different failure types. As shown in Figure~\ref{fig: class-wise-comparison}, SoloGen almost fails to remediate Pod Failure and Configuration Error issues, whereas ThinkRemed achieves a certain level of success in both cases. For other failure categories that both methods can handle, ThinkRemed consistently achieves higher remediation accuracy. It is also noteworthy that both methods exhibit very limited effectiveness in handling IO Saturation failures.

\section{Conclusion}

In this paper, we introduce MicroRemed, a benchmark designed to evaluate the end-to-end microservice remediation capabilities of LLMs. We also proposed ThinkRemed, a multi-agent framework that emulates the iterative decision-making process of SREs in microservice environments. Experimental results demonstrate that MicroRemed poses substantial challenges to existing LLMs, while ThinkRemed’s ability to perceive and reflect on system states enhances end-to-end remediation performance. Our work underscores the importance of achieving fully automated microservice remediation, paving the way toward more scalable and reliable LLM-driven software maintenance.

\section*{Limitations}

We discuss the limitations of our work from two perspectives: the benchmark and the methodology.

\noindent \textbf{Benchmark.}
Although the MicroRemed benchmark provides sufficient challenges for evaluating end-to-end microservice remediation, the currently supported failure types remain limited—covering only seven of the most common categories. In real-world systems, failure modes are far more diverse and continuously evolving~\cite{zhang2024multivariate, zhang2024towards, wang2025survey}. Nevertheless, the design of MicroRemed inherently supports extensibility; new failure types can be integrated seamlessly. The main challenge lies in the need to implement corresponding fault injection and detection mechanisms when introducing additional failure types.

\noindent \textbf{Methodology.}
While ThinkRemed, as an end-to-end microservice remediation framework, theoretically accesses all necessary runtime information and supports iterative reflection and reasoning, prior studies suggest that incorporating additional data—such as source code~\cite{pei2025flow, li2025coca} or historical remediation records~\cite{chen2024automatic, roy2024exploring}—can further enhance software maintenance. Moreover, building more sophisticated, domain-specific agent systems~\cite{zhang2025survey, yang2025swe} may also lead to improved performance. To this end, MicroRemed offers flexible interfaces to facilitate the integration of such advanced approaches in future work.

\balance
\bibliography{custom}

\newpage
\onecolumn
\appendix

\begin{center}
	{\huge \bfseries Appendices}
\end{center}

\vspace{3em}

\noindent{\Large \bfseries Table of Contents}

\vspace{0.5em}

\hrule height 0.5pt  

\startcontents[appendix]
\printcontents[appendix]{}{1}{}

\vspace{0.7em}

\hrule height 0.5pt  

\newpage

\section{Discussion}
\label{sec:discussion}

In this section, we discuss the potential applications and extensibility of the MicroRemed benchmark. As noted earlier, MicroRemed is inherently scalable. Although we provide two reference solutions—SoloGen and ThinkRemed—the benchmark is designed to accommodate future methods that may better address the end-to-end microservice remediation task. Researchers are encouraged to replace or extend the existing methods to explore new remediation strategies under the same evaluation framework.

The simplest way to use MicroRemed is to directly substitute the underlying LLM within the ThinkRemed framework. Since ThinkRemed enables access to dynamic runtime and environmental information and supports iterative reflection and reasoning, it provides a versatile interface for testing diverse LLM-based approaches. In theory, if the backbone LLM demonstrates sufficient reasoning and generalization capabilities, ThinkRemed can achieve fully automated end-to-end microservice remediation, highlighting the promise of LLM-driven system reliability and maintenance.

\section{Detailed Workflow of Software Maintenance}
\label{sec:workflow}

\begin{figure*}[hbp]
	\centering
	\includegraphics[width=1\linewidth]{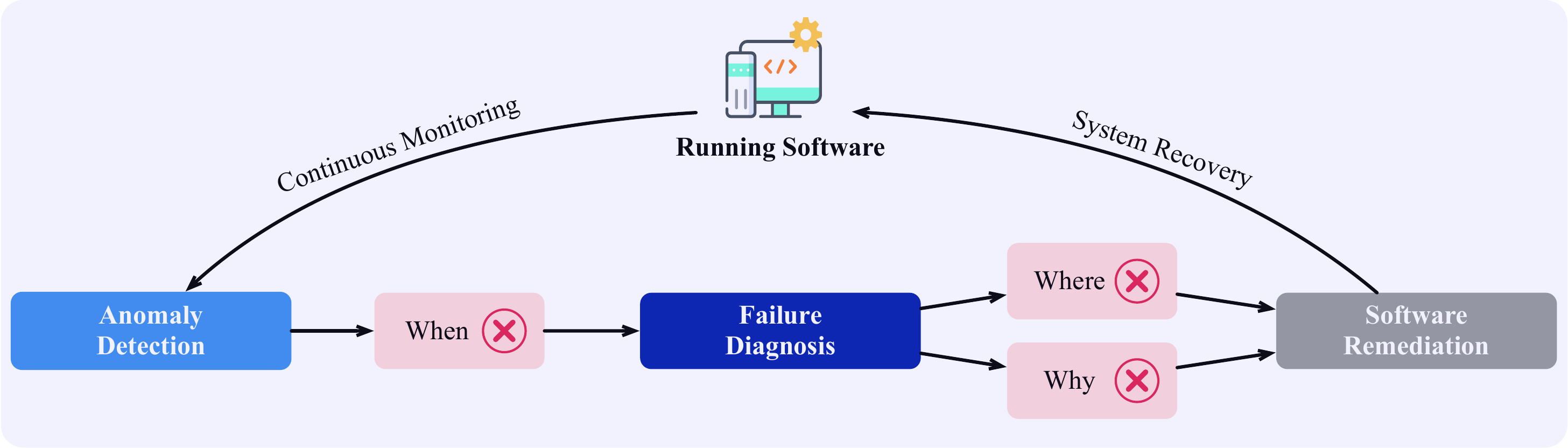}
	\caption{The overall workflow of software maintenance}
	\label{fig: fm-workflow}
\end{figure*}

The overall workflow of software maintenance is illustrated in Figure~\ref{fig: fm-workflow}. In a running software system, anomaly detection continuously monitors system behavior to identify potential failures. Once a failure occurs, failure diagnosis performs an in-depth analysis to determine where and why the failure happened. Software remediation, which is the focus of this paper, generates appropriate recovery plans or scripts based on the diagnostic results, ultimately achieving system recovery. The following sections provide detailed descriptions and formal definitions of each stage.

\subsection{Anomaly Detection}

Anomaly detection serves as the first stage in the software maintenance workflow, responsible for identifying deviations from normal system behavior during runtime. 
Formally, let the system state at time $t$ be represented as a multivariate observation vector $\mathbf{x}_t \in \mathbb{R}^d$, derived from various runtime data sources such as logs, metrics, and traces. The goal of anomaly detection is to learn a mapping $f_{\text{det}}: \mathbf{x}_t \mapsto y_t$, where $y_t \in \{0, 1\}$ indicates whether the current state is normal ($0$) or anomalous ($1$).

In practice, anomaly detection operates in a continuous monitoring manner, 
maintaining a sliding window of recent observations $\mathbf{X}_{t-w:t} = \{\mathbf{x}_{t-w+1}, \dots, \mathbf{x}_t\}$ and evaluating the probability of failure occurrence as Equation~\ref{eq: anomaly-detection}, where $\theta_{\text{det}}$ is a dynamically adjusted threshold. Once this condition is satisfied, a potential failure event is triggered for further diagnostic analysis.

\begin{equation}
	P(y_t = 1 \mid \mathbf{X}_{t-w:t}) > \theta_{\text{det}}
	\label{eq: anomaly-detection}
\end{equation}

Modern implementations often integrate statistical inference, unsupervised learning, 
and LLM-based semantic reasoning to detect both quantitative anomalies (e.g., metric deviations) and qualitative anomalies (e.g., abnormal log semantics).

\subsection{Failure Diagnosis}

Once the anomaly detection module identifies that the software system exhibits abnormal behavior within a specific time window, the process enters the \textit{Failure Diagnosis} stage. Failure diagnosis aims to localize the origin of the detected failure and determine the underlying cause of its occurrence. It is typically decomposed into two sequential sub-tasks: (1) \textbf{Failure Localization} (\textit{where}) and (2) \textbf{Failure Category Classification} (\textit{why}). Both stages operate on the detected anomaly event $e_t$ and the contextual information
$\mathcal{C}_t = \{\mathbf{X}_{t-w:t}, \mathcal{L}_t, \mathcal{T}_t\}$, where $\mathbf{X}_{t-w:t}$ denotes recent metric observations, $\mathcal{L}_t$ denotes system logs, and $\mathcal{T}_t$ denotes trace data.

\paragraph{(1) Failure Localization.}
The localization module identifies candidate faulty components (e.g., services, pods, or nodes) and ranks them by their likelihood of being the root cause. Formally, localization yields a ranked list, as illustrated in Equation~\ref{eq: failure-localization}, where each $r_i$ is a component identifier, and $q_i = P(r_i \mid e_t, \mathcal{C}_t)$ represents the estimated probability that component $r_i$ is responsible for the observed anomaly.

\begin{equation}
    \mathcal{R}_t^{\text{loc}} = \big\{(r_1, q_1), (r_2, q_2), \ldots, (r_m, q_m)\big\}
    \label{eq: failure-localization}
\end{equation}

The localization function can thus be defined as Equation~\ref{eq: localization-function}, where the ranking satisfies $q_1 \ge q_2 \ge \cdots \ge q_m$.

\begin{equation}
    \mathcal{R}_t^{\text{loc}} = f_{\text{loc}}(e_t, \mathcal{C}_t),
    \label{eq: localization-function}
\end{equation}

Practical localization techniques include statistical correlation analysis between metrics and components, trace-based dependency reasoning, and LLM-augmented inference that interprets logs and contextual signals to enhance localization accuracy.

\paragraph{(2) Failure Category Classification.}
Given a localized component $r_i$ (typically from the top-$k$ ranked candidates), the classification module predicts the failure category $c \in \mathcal{F}$ (e.g., \textit{CPU}, \textit{Memory}, \textit{I/O}, \textit{Network}, \textit{Configuration}). For each candidate, we compute as Equation~\ref{eq: failure-classification} and select the most probable class as Equation~\ref{eq: failure-classification-select}.

\begin{equation}
    P(c \mid r_i, e_t, \mathcal{C}_t) \quad \text{for } c \in \mathcal{F},
    \label{eq: failure-classification}
\end{equation}

\begin{equation}
    c_i^* = \arg\max_{c \in \mathcal{F}} P(c \mid r_i, e_t, \mathcal{C}_t)
    \label{eq: failure-classification-select}
\end{equation}

The classification output for each localized component $r_i$ is represented as a labeled pair $(r_i, c_i^*)$, accompanied by its confidence score $P(c_i^* \mid r_i, e_t, \mathcal{C}_t)$. The complete diagnostic result can thus be expressed as Equation~\ref{eq: failure-diagnosis}, where $p_i \triangleq P(c_i^* \mid r_i, e_t, \mathcal{C}_t)$ denotes the confidence of the predicted failure category for component $r_i$.

\begin{equation}
    \mathcal{R}_t = \big\{(r_1, c_1^*, p_1), (r_2, c_2^*, p_2), \ldots\big\}
    \label{eq: failure-diagnosis}
\end{equation}

\subsection{Software Remediation}

Based on the results of failure diagnosis, the system performs \textit{software remediation} sequentially according to the descending order of failure probabilities $p_i$. For each diagnosed failure tuple $(r_i, c_i^*)$, where $r_i$ denotes the localized faulty region and $c_i^*$ represents the predicted failure category, the remediation module initiates a targeted recovery process. 

Software remediation is an action-oriented stage that operationalizes diagnostic insights into concrete recovery strategies or executable repair scripts. The objective is to restore the system to a stable and healthy state with minimal service disruption. Formally, software remediation can be expressed as a mapping function shown in Equation~\ref{eq: software-remediation}, where $\mathcal{S}_t$ denotes the current system state and $\mathcal{A}_i$ represents the remediation action set. Each $\mathcal{A}_i$ may include actions such as service restart, configuration rollback, resource reallocation, or patch deployment.

\begin{equation}
    \mathcal{R}: (r_i, c_i^*, \mathcal{S}_t) \rightarrow \mathcal{A}_i,
    \label{eq: software-remediation}
\end{equation}

The effectiveness of remediation is continuously evaluated through the observed post-remediation system state $\mathcal{S}_{t+1}$, enabling a feedback loop for iterative refinement. The remediation process can thus be formalized as a closed-loop optimization, expressed as Equation~\ref{eq: software-remediation-evaluation}.

\begin{equation}
    \min_{\mathcal{A}_i} \; \mathbb{E}\big[\mathcal{L}(\mathcal{S}_{t+1})\big],
    \label{eq: software-remediation-evaluation}
\end{equation}

Among them, $\mathcal{L}(\cdot)$ denotes the system loss function quantifying deviation from normal operation. A successful remediation satisfies the condition $\mathcal{L}(\mathcal{S}_{t+1}) \leq \delta$, where $\delta$ is a predefined threshold for acceptable system stability.

In practice, remediation strategies can be either \textit{plan-based}, in which a recovery plan is synthesized through rule-based or policy-driven reasoning, or \textit{script-based}, where executable scripts are generated to automate the recovery process. In this paper, our focus is on the latter—leveraging LLM-driven reasoning to autonomously generate and validate repair scripts that achieve full end-to-end system recovery.

\section{Detailed Specification of Ansible Playbook}
\label{sec:ansible-playbook}

In this section, we provide a detailed description of Ansible and its interaction with microservice systems. We first explain how Ansible fits into the management of distributed microservice environments, and then provide concrete examples of Ansible Playbooks to illustrate their practical use.

\subsection{Microservice System \& Ansible}

Modern software systems are increasingly built upon the microservice architecture, where a complex application is decomposed into a collection of small, independently deployable services. Each service is typically containerized, deployed across multiple nodes, and interconnected through APIs or message queues. This design offers scalability and flexibility, yet also introduces operational complexity due to dynamic dependencies, frequent updates, and heterogeneity in the runtime environment.

To manage such distributed systems effectively, automation becomes essential. Manual operations, such as configuration updates, dependency installation, and service restarts, are error-prone and infeasible at scale. This is where \textbf{Ansible} plays a crucial role. As an open-source automation framework, Ansible provides a declarative and agentless approach to orchestrate and control distributed resources.

Ansible operates by connecting to remote hosts via SSH or API interfaces and executing predefined instructions, called \textit{tasks}, written in YAML-based Playbooks. Unlike imperative scripting languages (e.g., Bash or Python scripts), Ansible adopts a \textit{declarative model} that specifies the desired state of the system rather than the step-by-step commands to achieve it. This design brings several significant benefits:

\begin{itemize}
    \item \textbf{Consistency and Idempotence:} Repeated executions converge the system to a stable and predictable state, avoiding redundant or conflicting operations.
    \item \textbf{Abstraction and Modularity:} Tasks can be grouped into roles and modules, enabling the reuse of operational logic across multiple services.
    \item \textbf{Agentless Deployment:} No additional software needs to be installed on managed nodes, simplifying integration with existing infrastructure.
    \item \textbf{Scalability:} Through inventory configuration, Ansible can manage hundreds or thousands of service instances simultaneously.
\end{itemize}

In microservice environments, Ansible serves as a unifying layer that bridges the management of heterogeneous resources—such as containers, virtual machines, and network configurations—under a single, coherent control model. It enables operators to define, apply, and verify system states in a structured and auditable manner, which is particularly valuable for continuous deployment, configuration synchronization, and system maintenance.

\subsection{Ansible Playbook Examples}

An Ansible Playbook is a YAML-based declarative script that defines a sequence of automation tasks to be executed on one or more target systems. It acts as a “script” that instructs Ansible what operations to perform on which target hosts in order to achieve automation goals such as configuration management, application deployment, and system maintenance.Each playbook describes a set of \textit{hosts}, \textit{tasks}, and their corresponding \textit{actions}, allowing administrators or autonomous agents to specify \textit{what} to do rather than \textit{how} to do it. Compared with ad-hoc shell scripts, playbooks provide higher-level abstractions with clear semantics, reusability, and idempotence, making them particularly suitable for large-scale automated system maintenance.

To illustrate the role of an ansible playbook in microservice remediation, consider a simple example addressing a high CPU load issue. When a monitoring system detects that the CPU utilization of a given service instance exceeds a predefined threshold, 
the remediation logic can automatically trigger a scaling operation to relieve the overload. The corresponding playbook may be written as Figure~\ref{fig: ansible-playbook}.

\lstset{
  basicstyle=\ttfamily\footnotesize, 
  backgroundcolor=\color{gray!10},
  frame=single,
  rulecolor=\color{black},
  frameround=tttt,
  showstringspaces=false,
  breaklines=true,
  postbreak=\mbox{\textcolor{red}{$\hookrightarrow$}\space},
  numbers=left,                     
  numberstyle=\tiny\color{gray},    
  xleftmargin=2em,                  
  xrightmargin=2em,                 
  keywordstyle=\color{blue},        
  stringstyle=\color{teal},         
}
\begin{figure}[htbp]
\centering
\begin{lstlisting}
---
- name: Mitigate high CPU load by scaling service replicas
  hosts: microservice_nodes
  become: yes
  tasks:
    - name: Check current CPU utilization
      shell: "top -bn1 | grep 'Cpu(s)' | awk '{print $2 + $4}'"
      register: cpu_load

    - name: Scale service if CPU utilization exceeds 80%
      shell: "kubectl scale deployment my-service --replicas=4"
      when: cpu_load.stdout | float > 80.0

    - name: Notify monitoring system
      shell: "curl -X POST http://monitor/api/notify -d 'scale-up executed'"
\end{lstlisting}
\caption{An Ansible Playbook for CPU scaling operation}
\label{fig:ansible-playbook}
\end{figure}

This example demonstrates how Ansible bridges the gap between diagnosis and action: it performs real-time monitoring, conditional execution, and service orchestration within a single declarative workflow. In real-world microservice systems, such playbooks can be dynamically generated by LLM-based agents according to diagnostic results (e.g., faulty component and failure category), enabling fully automated and adaptive system remediation.

\section{Coordinator Prompts}
\label{sec:coordinator-prompts}

In this section, we present several core prompts used in the Coordinator of ThinkRemed. Note that these are not exhaustive — the complete set of prompts can be found in the released source code.

As illustrated in Figure~\ref{fig:role-definition-prompt}, the \textit{Role Definition Prompt} serves as the system-level initialization prompt, guiding the model to act as an experienced Site Reliability Engineer (SRE) responsible for orchestrating the entire remediation process. It defines the execution environment, provides diagnostic information, and instructs the model to generate an executable Ansible playbook for fault recovery.

Following remediation attempts, ThinkRemed employs a reflective prompting mechanism to handle unsuccessful recovery cases. As shown in Figure~\ref{fig:regeneration-prompt}, the \textit{Regeneration Prompt} instructs the model to reason about the cause of the previous failure, leverage additional probe agents to gather more runtime evidence, and subsequently regenerate a refined playbook for another recovery attempt. This iterative prompting process enables adaptive and autonomous remediation across complex runtime conditions.

In addition to these examples, ThinkRemed contains a variety of other prompts for tool invocation, failure context summarization, and strategy selection, which are all detailed in the released source code repository.

\begin{figure}[h]
	\centering
	\begin{tcolorbox}[colback=gray!10,
		colframe=black,
		width=\linewidth,
		arc=1mm, auto outer arc,
		boxrule=0.5pt,
		top=2pt, bottom=2pt, left=2pt, right=2pt]
		You are an experienced SRE managing a microservice system.\\
        A failure has occurred, and your task is to generate a final executable Ansible playbook based on the given root cause, failure category, and the probed information (executed by ``ansible-playbook -i inventory.ini remediation.yml'').\\
        The system will automatically execute the playbook and verify whether the failure has been successfully resolved.\\
        {[}Attention{]} Please ensure that online services remain uninterrupted; restarting services should not be considered a primary strategy.\\
        \{runtime\_envs\}\\
        The content of inventory.ini is \{inventory\_content\}\\
        The current namespace is: \{namespace\}, failure root cause service is: \{root\_cause\}, and the failure category is: \{failure\_category\}.
	\end{tcolorbox}
	\caption{The Prompt for Role Definition}
	\label{fig:role-definition-prompt}
\end{figure}

\begin{figure}[h]
	\centering
	\begin{tcolorbox}[colback=gray!10,
		colframe=black,
		width=\linewidth,
		arc=1mm, auto outer arc,
		boxrule=0.5pt,
		top=2pt, bottom=2pt, left=2pt, right=2pt]
		The failure of online service has not yet been remediated.\\
        You may use the probe agent to further inspect the system state and generate a new Ansible playbook to attempt remediation again.\\
        The previous playbook execution returned: \{playbook\_exec\_status\}, output: \{status\}
	\end{tcolorbox}
	\caption{The Prompt for Regeneration}
	\label{fig:regeneration-prompt}
\end{figure}

\section{Case Study}
\label{sec:case-study}

We present a practical example of using \textbf{ThinkRemed} for microservice remediation. In this case, a \textbf{CPU Saturation} failure was injected into the \texttt{ts-news-service} pod of the \textbf{Train-Ticket} microservice system.

ThinkRemed first attempted to collect environment information by generating the following command sequence: 
\texttt{kubectl get deployment ts-news-service -n train-ticket -o jsonpath=\{.spec.replicas\}; kubectl get deployment ts-news-service -n train-ticket -o jsonpath=\{.spec.template.spec.containers[0].resources.limits.cpu\}; kubectl describe hpa ts-news-service -n train-ticket 2>/dev/null || echo 'No HPA found'}. 
The corresponding responses were \texttt{"1; 500m; No HPA found"}. 

\begin{figure}[hbp]
\centering
\begin{lstlisting}
---
- name: Remediate CPU stress on ts-news-service
  hosts: k3s_control_plane
  tasks:
    - name: Scale deployment to increase replicas
      command: kubectl scale deployment ts-news-service --namespace=train-ticket --replicas=3
      args:
        executable: /bin/bash
      register: scale_result
      ignore_errors: yes
      changed_when: "'scaled to' in scale_result.stderr"
    - name: Verify deployment scale
      command: kubectl get deployment ts-news-service -n train-ticket -o jsonpath='{.spec.replicas}'
      register: verify_result
      failed_when: verify_result.stdout | int < 3
\end{lstlisting}
\caption{Ansible Playbook Produced After the First Iteration of Reasoning}
\label{fig:case-study-playbook1}
\end{figure}

Then, ThinkRemed reasoned that additional information might be helpful and issued another command: 
\texttt{kubectl get deployment ts-news-service -n online-boutique -o jsonpath=\{.spec.template.spec.containers[0].resources.requests.cpu\}}, 
which returned \texttt{"100m"}. 

After this step, ThinkRemed concluded that it had gathered sufficient information and subsequently generated the playbook shown in Figure~\ref{fig:case-study-playbook1}.

The Execution Agent automatically executed the generated playbook. However, verification by the Verification Agent revealed that the issue remained unresolved. This was because the injected \textbf{CPU Saturation} failure was precisely targeted at a specific pod, meaning that simply increasing the number of replicas could not mitigate the problem. Consequently, ThinkRemed was prompted to perform reflection and regeneration, eventually producing the Ansible playbook shown in Figure~\ref{fig:case-study-playbook2}.

\begin{figure}[hbp]
\centering
\begin{lstlisting}
---
- name: Remediate CPU stress via resource adjustment
  hosts: k3s_control_plane
  tasks:
    - name: Increase CPU limits for ts-news-service pods
      command: kubectl set resources deployment/ts-news-service -n train-ticket --requests=cpu=500m --limits=cpu=1000m --record
      args:
        executable: /bin/bash
      register: resource_update
      ignore_errors: yes
    - name: Trigger deployment rollout
      command: kubectl rollout restart deployment/ts-news-service -n train-ticket
      args:
        executable: /bin/bash
      register: rollout_result
      ignore_errors: yes
    - name: Verify CPU resource allocation
      command: kubectl get deployment ts-news-service -n train-ticket -o jsonpath='{.spec.template.spec.containers[0].resources.limits.cpu}'
      register: verify_limit
      failed_when: verify_limit.stdout != '1000m'
\end{lstlisting}
\caption{Ansible Playbook Produced After Reflection}
\label{fig:case-study-playbook2}
\end{figure}

This playbook mitigates the performance bottleneck or excessive CPU pressure caused by insufficient CPU resources by increasing the CPU quota of the ts-news-service pods and enforcing a rolling update. Specifically, it raises the CPU limit from \texttt{500m} to \texttt{1000m}, thereby eliminating the chaos-injected \textbf{CPU Saturation} failure and successfully restoring the system to a healthy state.

\section{Experiment Results for Ablation Study}
\label{sec:ablation-study}

To evaluate the effectiveness of each component in ThinkRemed, we conducted an ablation study using Qwen3-Plus, which demonstrated the best performance among all models tested in the MicroRemed benchmark across three microservice systems.

\begin{table*}[hbp]
	\setlength{\tabcolsep}{6pt}
	\centering
	\begin{tabular}{c|ccc|ccc|ccc}
		\toprule
		Method & \multicolumn{3}{c|}{Train-Ticket} & \multicolumn{3}{c|}{Online-Boutique} & \multicolumn{3}{c}{Simple-Micro} \\
		\\[-2.5ex]
		\cline{2-10}
		\\[-2ex]
		~ & \textit{Easy} & \textit{Medium} & \textit{Hard} & \textit{Easy} & \textit{Medium} & \textit{Hard} & \textit{Easy} & \textit{Medium} & \textit{Hard} \\
		\midrule
		\textbf{ThinkRemed} & \textbf{47.83} & 30.61 & \textbf{31.17} & \textbf{43.48} & \textbf{43.75} & \textbf{31.58} & \textbf{47.83} & 36.17 & \textbf{38.03} \\
        \midrule
		w/o Probe & \underline{43.48} & \textbf{34.69} & \underline{30.38} & \underline{39.13} & \underline{40.43} & \underline{30.38} & \underline{43.48} & \textbf{38.78} & \underline{36.36} \\
		w/o Reflection & \underline{43.48} & 28.57 & 26.92 & 34.78 & 36.17 & 25.32 & 34.78 & 28.57 & 28.38 \\
        w/o P. \& R. & 39.13 & \underline{33.33} & 20.51 & 30.43 & 35.42 & 20.51 & 30.43 & \underline{36.73} & 26.15 \\
		\bottomrule
	\end{tabular}
	\caption{Ablation study with Qwen3-Plus as backbone}
    \label{tab:ablation-study}
\end{table*}

As shown in Table~\ref{tab:ablation-study}, the results are reported for three settings: without the probe agent, without reflection, and without both probe agent and reflection. Notably, removing both components effectively degenerates the system into SoloGen.

In most cases, both the probe agent and reflection mechanisms contribute positively to performance. For instance, in the Train-Ticket microservice (easy level), removing either the probe agent or reflection leads to a 13.05\% drop in accuracy. However, overall, reflection has a greater impact than the probe agent — on average, removing reflection results in a 7.16\% decrease in accuracy, whereas removing the probe agent only leads to a 1.57\% decrease. Interestingly, in certain cases (e.g., the medium level of Train-Ticket and Simple-Micro), accuracy slightly improves when the probe agent is removed. Upon closer analysis, we found that this occurs because current models still have limited contextual reasoning ability — excessive probing may introduce noise and mislead the final Ansible playbook generation.

\section{Experiment Results for Remediation Latency}
\label{sec:more-remediation-latency}

We reports the Average Remediation Latency (ARL) measured in seconds across nine large language models under two methods, SoloGen and ThinkRemed, for three microservice systems and three difficulty levels in detail. As illustrated in Table~\ref{tab: more-remediation-latency}, the reported latency reflects the end-to-end time required to complete one remediation cycle, including the time for model reasoning, system probing, action execution, and verification of service recovery.

\begin{table*}[h]
	\setlength{\tabcolsep}{1.3pt}
	\centering
	\begin{tabular}{c|c|ccc|ccc|ccc}
		\toprule
		\multirow{2}{*}{\textbf{Backbone}} & \multirow{2}{*}{\textbf{Method}} & \multicolumn{3}{c|}{Train-Ticket} & \multicolumn{3}{c|}{Online-Boutique} & \multicolumn{3}{c}{Simple-Micro} \\
		\\[-2.5ex]
		\cline{3-11}
		\\[-2ex]
		~ & ~ & \textit{Easy} & \textit{Medium} & \textit{Hard} & \textit{Easy} & \textit{Medium} & \textit{Hard} & \textit{Easy} & \textit{Medium} & \textit{Hard} \\
		\midrule
		\multicolumn{11}{c}{\cellcolor{brightlavender!15} \textbf{\textit{Closed-Sourced LLMs}}} \\
		\midrule
		\multirow{2}{*}{Qwen3-Plus} & SoloGen & 40.49 & 34.17 & 38.87 & 31.41 & 32.48 & 44.19 & 46.13 & 44.42 & 42.34 \\
		~ & ThinkRemed & 79.83 & 77.44 & 81.22 & 53.35 & 83.77 & 59.35 & 71.26 & 75.79 & 78.18 \\
		\midrule
		\multirow{2}{*}{Qwen3-Max}
		~ & SoloGen & 43.35 & 39.76 & 10.37 & 28.20 & 24.59 & 41.79 & 26.67 & 21.21 & 36.19 \\
		~ & ThinkRemed & 69.87 & 86.38 & 12.36 & 34.82 & 46.49 & 48.20 & 37.02 & 39.73 & 52.43 \\
		\midrule
		\multirow{2}{*}{Qwen3-Flash}
		~ & SoloGen & 20.94 & 21.93 & 35.61 & 25.43 & 24.96 & 57.29 & 21.78 & 19.43 & 20.36 \\
		~ & ThinkRemed & 43.24 & 84.96 & 290.66 & 42.33 & 77.26 & 98.64 & 50.52 & 61.12 & 65.33 \\
		\midrule			
		\multicolumn{11}{c}{\cellcolor{capri!15} \textbf{\textit{Open-Sourced LLMs}}} \\
		\midrule
		\multirow{2}{*}{QwQ-32B}
		~ & SoloGen & 57.18 & 75.28 & 73.60 & 68.50 & 64.31 & 79.14 & 64.08 & 67.29 & 75.39 \\
		~ & ThinkRemed & 157.39 & 194.07 & 183.21 & 109.89 & 137.33 & 155.50 & 141.54 & 188.69 & 195.67 \\			
		\midrule	
		\multirow{2}{*}{Qwen3-Next}
		~ & SoloGen & 16.12 & 14.65 & 16.58 & 14.78 & 15.53 & 20.71 & 16.62 & 26.25 & 24.60 \\
		~ & ThinkRemed & 23.33 & 20.41 & 29.76 & 22.57 & 22.73 & 26.64 & 24.83 & 34.58 & 32.83 \\
		\midrule			
		\multirow{2}{*}{Qwen3-235B}
		~ & SoloGen & 39.60 & 54.52 & 38.81 & 26.33 & 43.79 & 33.52 & 31.06 & 32.68 & 49.79 \\
		~ & ThinkRemed & 83.34 & 92.20 & 73.54 & 74.57 & 55.52 & 74.27 & 82.49 & 66.65 & 73.20 \\					
		\midrule				
		\multirow{2}{*}{DeepSeek-V3.2}
		~ & SoloGen & 96.65 & 84.30 & 115.36 & 63.06 & 63.14 & 60.07 & 47.99 & 57.01 & 68.00 \\
		~ & ThinkRemed & 148.32 & 155.10 & 121.52 & 149.14 & 113.55 & 117.73 & 129.61 & 106.41 & 98.64 \\
		\midrule	
		\multirow{2}{*}{Kimi-K2}
		~ & SoloGen & 94.28 & 103.04 & 84.50 & 66.83 & 76.02 & 79.10 & 84.65 & 74.07 & 61.06 \\
		~ & ThinkRemed & 90.84 & 81.56 & 101.08 & 75.87 & 79.07 & 83.85 & 90.82 & 79.28 & 76.84 \\
		\midrule
		\multirow{2}{*}{GLM-4.5}
		~ & SoloGen & 72.11 & 76.26 & 48.12 & 61.60 & 69.26 & 66.97 & 64.17 & 42.91 & 39.50 \\
		~ & ThinkRemed & 189.21 & 112.07 & 108.04 & 135.10 & 126.03 & 130.82 & 127.82 & 126.26 & 132.82 \\
		\bottomrule
	\end{tabular}
	\caption{Average Remediation Latency (\textit{ARL}) per remediation, measured in seconds, across closed-source and open-source LLM backbones}
    \label{tab: more-remediation-latency}
\end{table*}

Overall, the results indicate that ThinkRemed consistently incurs higher remediation latency than SoloGen. This is expected, as ThinkRemed introduces additional procedural steps such as probing, iterative reflection, and verification before finalizing remediation decisions. These steps enable the model to gather more contextual evidence and perform reasoning-based correction but naturally extend the total execution time. Among the closed-source models, Qwen3-Plus demonstrates a well-balanced performance, achieving relatively moderate latency despite its complex reasoning steps. In contrast, Qwen3-Flash and Qwen3-Max exhibit large latency variances, with several extreme outliers (e.g., 290.66 seconds in Train-Ticket Hard), suggesting that their internal reasoning or retry mechanisms may occasionally lead to prolonged inference or repeated plan generation.

For open-source backbones, a similar trend is observed. Models such as QwQ-32B and GLM-4.5 show considerably higher latencies, even though their parameter scales are smaller. This phenomenon can be attributed to their forced multi-step reasoning mechanisms, which prolong the inference process without yielding proportionally higher accuracy. Conversely, Qwen3-Next achieves remarkably low latency across all environments, reflecting a lightweight reasoning path and efficient prompt processing. Nevertheless, such fast execution typically comes at the cost of reduced reasoning depth and less stable accuracy, indicating a trade-off between response efficiency and decision reliability.

The latency also varies significantly across microservice systems and difficulty levels. The Train-Ticket system, which has the most complex dependency graph and failure scenarios, consistently shows the highest ARL, whereas Online-Boutique and Simple-Micro display relatively shorter remediation times. Interestingly, while the “Hard” scenarios generally lead to longer latencies, several exceptions exist where the latency unexpectedly decreases. These cases are often associated with early termination of remediation due to timeouts or premature success detection in system verification.

Overall, the results highlight the fundamental trade-off between accuracy and latency in reasoning-based remediation. ThinkRemed improves robustness and decision reliability but incurs additional latency from iterative reasoning and verification. Models such as Qwen3-Plus achieve a favorable balance, maintaining high accuracy with moderate response time, whereas QwQ-32B and GLM-4.5 demonstrate that excessive deliberation may degrade time efficiency without significant accuracy gains. These findings suggest that future research should focus on optimizing the orchestration process—such as adaptive probing strategies, dynamic timeout adjustment, and selective reflection—to retain the benefits of reasoning-driven remediation while reducing overall latency.

\section{Experiment Results for Token Consumption}
\label{sec:more-token-consumption}

To further assess the efficiency of ThinkRemed, we analyze both its token consumption and remediation latency. Table~\ref{tab: more-token-consumption} summarizes the average token consumption (\textit{ATC}) per remediation across nine LLM backbones and three microservice systems under different difficulty levels. The reported numbers include both input and output tokens, thus reflecting the overall reasoning and generation workload required for each remediation process.

\begin{table*}[h]
	\setlength{\tabcolsep}{1.3pt}
	\centering
	\begin{tabular}{c|c|ccc|ccc|ccc}
		\toprule
		\multirow{2}{*}{\textbf{Backbone}} & \multirow{2}{*}{\textbf{Method}} & \multicolumn{3}{c|}{Train-Ticket} & \multicolumn{3}{c|}{Online-Boutique} & \multicolumn{3}{c}{Simple-Micro}\\
		\\[-2.5ex]
		\cline{3-11}
		\\[-2ex]
		~ & ~ & \textit{Easy} & \textit{Medium} & \textit{Hard} & \textit{Easy} & \textit{Medium} & \textit{Hard} & \textit{Easy} & \textit{Medium} & \textit{Hard}\\
		\midrule
		\multicolumn{11}{c}{\cellcolor{brightlavender!15} \textbf{\textit{Closed-Sourced LLMs}}} \\
		\midrule
		\multirow{2}{*}{Qwen3-Plus} & SoloGen & 768 & 765 & 728 & 745 & 678 & 758 & 716 & 760 & 725\\
		~ & ThinkRemed & 3363 & 4988 & 4359 & 18822 & 127299 & 108053 & 3863 & 5013 & 5225\\
		\midrule
		\multirow{2}{*}{Qwen3-Max}
		~ & SoloGen & 698 & 651 & 209 & 604 & 529 & 631 & 565 & 549 & 635\\
		~ & ThinkRemed & 3583 & 2145 & 256 & 4821 & 5366 & 5758 & 4405 & 3918 & 5454\\
		\midrule
		\multirow{2}{*}{Qwen3-Flash}
		~ & SoloGen & 979 & 982 & 995 & 1044 & 1023 & 978 & 986 & 1003 & 1042\\
		~ & ThinkRemed & 2948 & 3651 & 3891 & 3490 & 3003 & 3378 & 2057 & 2522 & 2679\\
		\midrule			
		\multicolumn{11}{c}{\cellcolor{capri!15} \textbf{\textit{Open-Sourced LLMs}}} \\
		\midrule
		\multirow{2}{*}{QwQ-32B}
		~ & SoloGen & 489 & 436 & 479 & 427 & 418 & 457 & 455 & 436 & 451\\
		~ & ThinkRemed & 4147 & 5918 & 5936 & 2003 & 3784 & 2951 & 3449 & 3272 & 3782\\			
		\midrule	
		\multirow{2}{*}{Qwen3-Next}
		~ & SoloGen & 742 & 771 & 791 & 778 & 728 & 799 & 776 & 763 & 815\\
		~ & ThinkRemed & 14490 & 13267 & 13091 & 12190 & 9387 & 24385 & 7060 & 7820 & 9549\\
		\midrule			
		\multirow{2}{*}{Qwen3-235B}
		~ & SoloGen & 691 & 705 & 688 & 713 & 678 & 748 & 814 & 692 & 752\\
		~ & ThinkRemed & 4858 & 8369 & 6669 & 10624 & 15749 & 30381 & 5527 & 6497 & 6078\\					
		\midrule				
		\multirow{2}{*}{DeepSeek-V3.2}
		~ & SoloGen & 544 & 580 & 625 & 595 & 625 & 636 & 615 & 603 & 600\\
		~ & ThinkRemed & 5195 & 7295 & 7785 & 5855 & 6004 & 6454 & 6341 & 6581 & 6454 \\
		\midrule	
		\multirow{2}{*}{Kimi-K2}
		~ & SoloGen & 1122 & 1174 & 1227 & 1085 & 1133 & 1182 & 1079 & 1124 & 1186\\
		~ & ThinkRemed & 6452 & 4964 & 6728 & 4810 & 6314 & 7793 & 4022 & 7176 & 6874\\
		\midrule
		\multirow{2}{*}{GLM-4.5}
		~ & SoloGen & 407 & 409 & 378 & 524 & 472 & 394 & 547 & 449 & 430\\
		~ & ThinkRemed & 11264 & 9492 & 10652 & 11270 & 11991 & 10692 & 7910 & 10265 & 9750\\
		\bottomrule
	\end{tabular}
	\caption{Average Token Consumption (\textit{ATC}) per remediation across closed-source and open-source LLM backbones}
    \label{tab: more-token-consumption}
\end{table*}

As shown, ThinkRemed consistently consumes significantly more tokens than SoloGen across all backbones and settings. This overhead primarily arises from ThinkRemed’s agentic reasoning mechanism, which performs iterative probing, reflection, and verification before generating the final remediation playbook. For instance, in the case of Qwen3-Plus—the strongest model in the benchmark—token usage increases from several hundred tokens under SoloGen to more than 100K tokens in hard-level scenarios of the Online-Boutique system. Similar patterns can be observed in other models such as Qwen3-235B and GLM-4.5, where the multi-turn reasoning process expands the interaction context and hence the token footprint.

When examined jointly with the Average Remediation Latency (ARL) results, a consistent trend emerges: both ATC and ARL grow proportionally with task difficulty and system complexity. The increase in latency is largely a consequence of the longer reasoning trajectories that ThinkRemed initiates to ensure correctness. Nevertheless, these additional computational costs are accompanied by clear performance benefits. Compared with SoloGen, ThinkRemed achieves an average improvement of over 7\% in remediation accuracy, indicating that the extra reasoning depth—though expensive in terms of tokens and time—enhances the model’s ability to interpret system failures and generate actionable playbooks.

Overall, the results highlight an explicit trade-off between reasoning depth and efficiency. ThinkRemed demonstrates that multi-agent, reflective reasoning substantially improves recovery success rates, albeit at the expense of higher token consumption and latency. This suggests that future optimization efforts could focus on reducing redundant reasoning steps or compressing intermediate reflections, aiming to retain accuracy gains while mitigating the token and time overhead inherent to agentic LLM workflows.

\section{Experiment Results for Hyperparameter Evaluation}
\label{sec:hyperparameter}

\begin{figure}[hbp]
			\centering
			\subfigure[Train-Ticket]{
				\begin{minipage}{0.31\linewidth}
					\centering   
					\includegraphics[width=\textwidth]{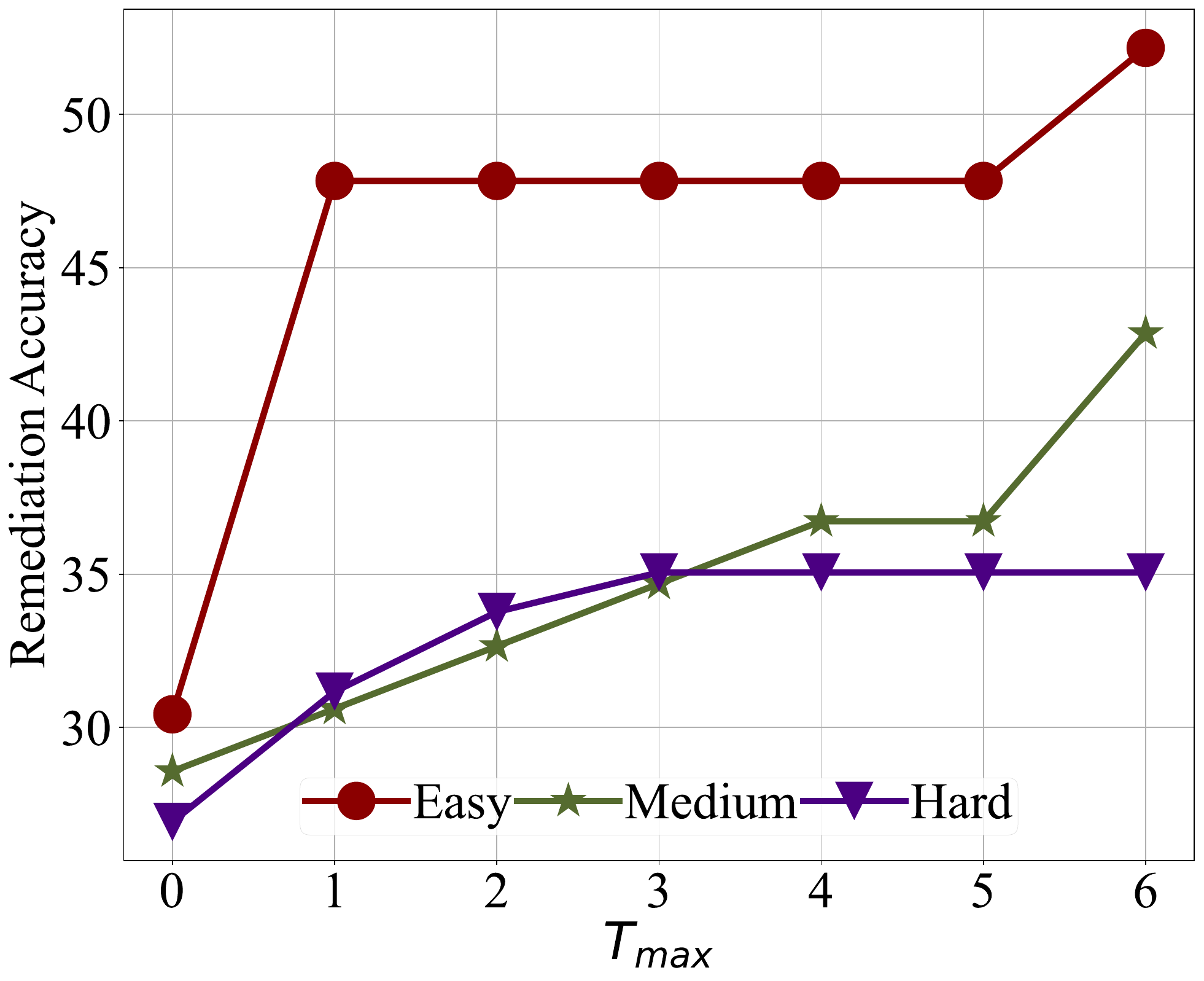}
					\label{fig: hyperparameter_train-ticket}
				\end{minipage}
			}
			\subfigure[Online-Boutique]{
				\begin{minipage}{0.31\linewidth}
					\centering
					\includegraphics[width=\textwidth]{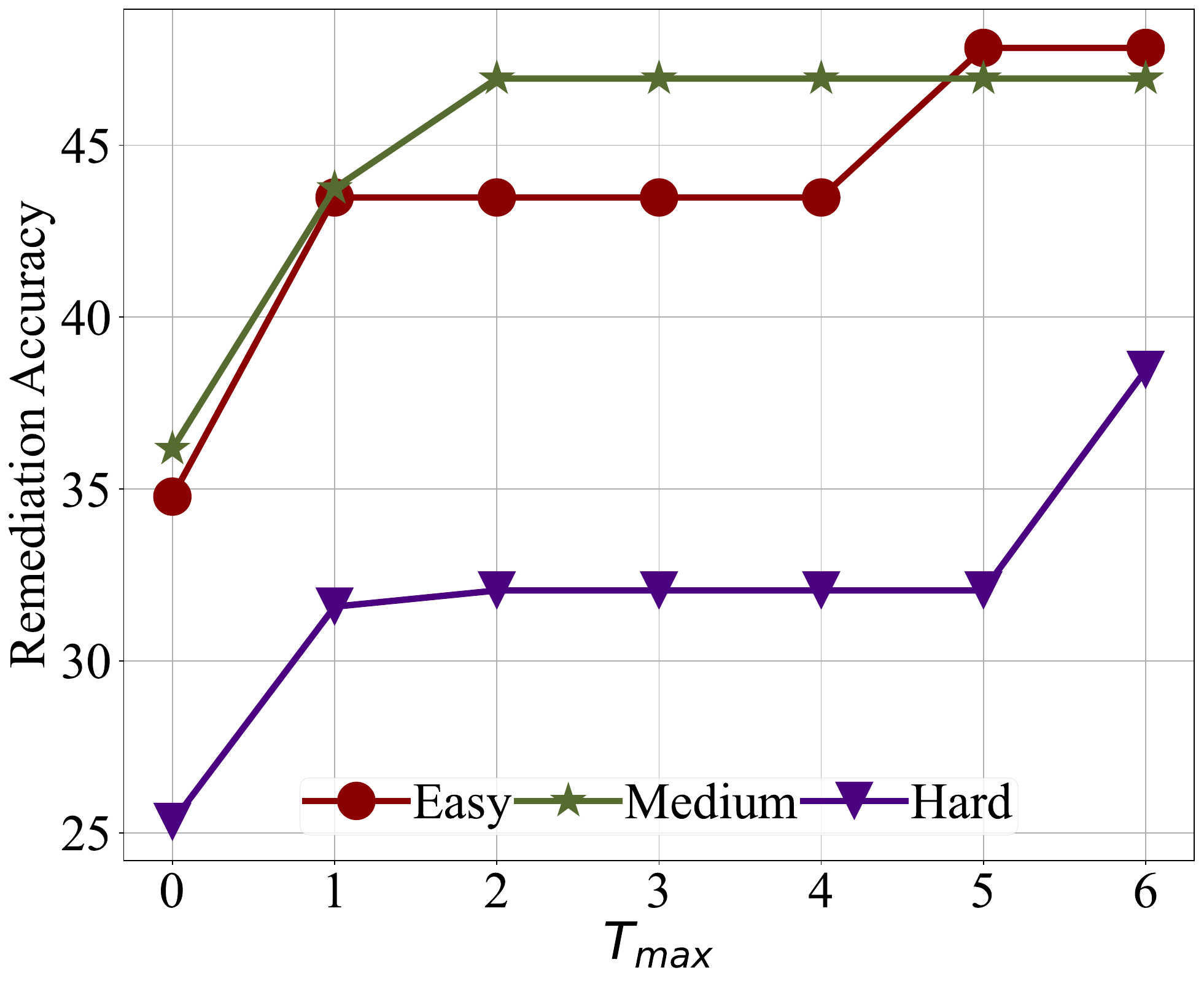}
					\label{fig: hyperparameter_online-boutique}
				\end{minipage}
			}
            \subfigure[Simple-Micro]{
				\begin{minipage}{0.31\linewidth}
					\centering
					\includegraphics[width=\textwidth]{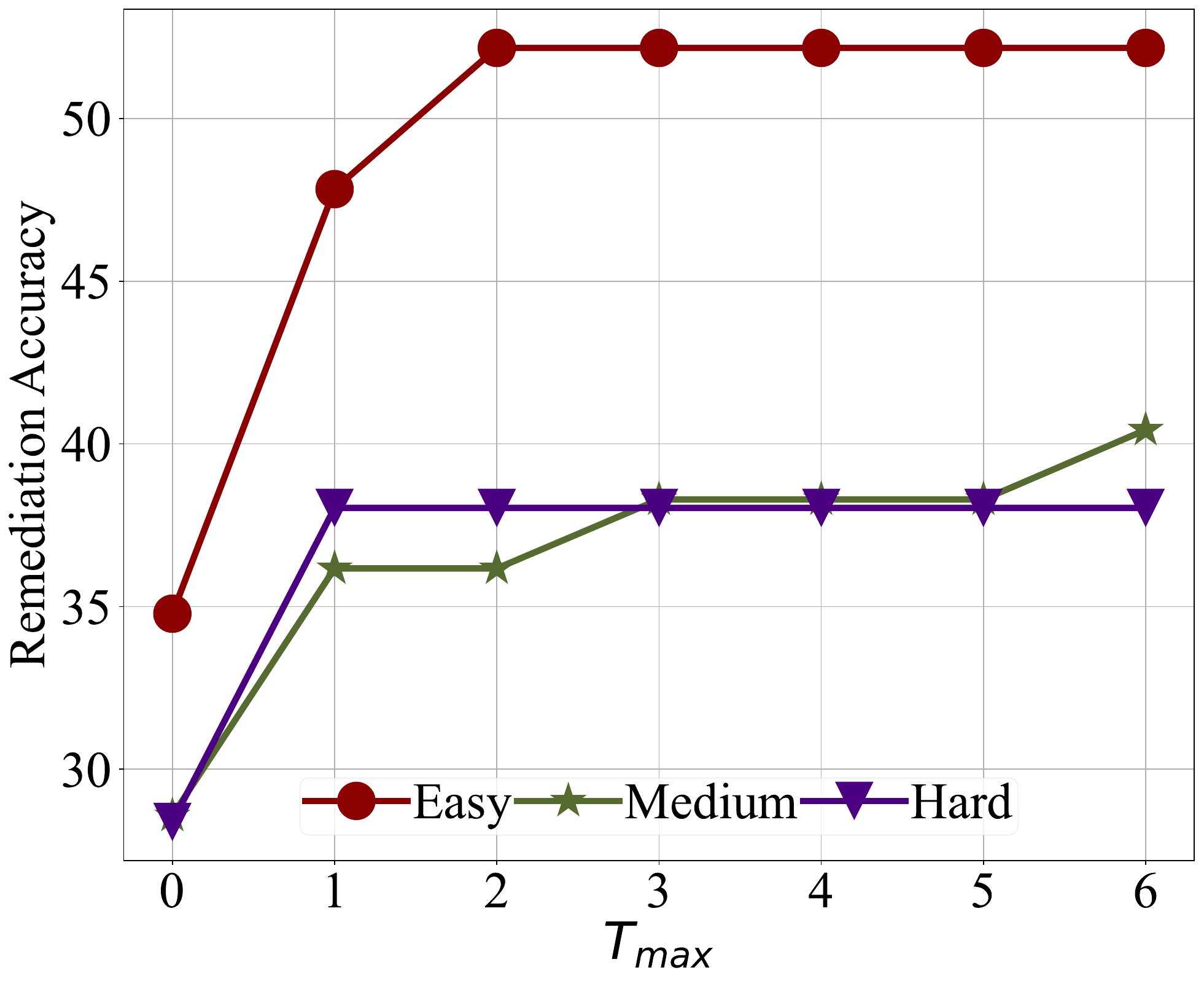}
					\label{fig: hyperparameter_simple-micro}
				\end{minipage}
			}
			\caption{Hyperparameter results with different values of $T_{max}$ (maximum retry count)}
			\label{fig: hyperparameter}
\end{figure}

To evaluate the effect of the reflection depth in \textsc{ThinkRemed}, we conduct hyperparameter experiments by varying the maximum reflection count $T_{max}$. This parameter controls how many reasoning-reflection iterations the system performs before generating the final remediation plan. Figure~\ref{fig: hyperparameter} illustrates the results across the three microservice systems—Train-Ticket, Online-Boutique, and Simple-Micro—under easy, medium, and hard task levels.

Overall, remediation accuracy shows a clear upward trend as $T_{max}$ increases from 0 to 6, but with diminishing returns beyond a certain point. For the Train-Ticket system, accuracy improves rapidly from 30.43\% to 52.17\% at the easy level as $T_{max}$ grows, while in the medium and hard levels, the growth is more gradual. A similar pattern is observed in Online-Boutique, where accuracy initially increases sharply (e.g., from 34.78\% to 47.83\% for easy) and then stabilizes, indicating that excessive reflections do not necessarily lead to further improvements.

The Simple-Micro system exhibits the most consistent benefit from higher $T_{max}$, particularly at the easy level, where accuracy reaches 52.17\% with $T_{max} \ge 2$. However, in more difficult settings, performance gains plateau after $T_{max}=3$, implying that additional reflection steps contribute marginally to reasoning quality.

These results suggest that reflection depth is indeed a crucial factor in improving remediation accuracy, especially when the reasoning space is moderately complex. However, the marginal gains diminish after a moderate number of reflections, likely because the LLM’s internal representation becomes saturated or redundant reasoning loops fail to introduce new insights. Considering both the performance gains and the corresponding increases in token consumption and latency (as shown in Section~\ref{sec:more-token-consumption}), setting $T_{max}$ within a small range provides a practical trade-off between accuracy and efficiency for most microservice scenarios.

\section{Experiment Results for Failure-Type-Wise Evaluation}
\label{sec:class-wise}

\begin{figure*}[hbp]
	\centering
	\includegraphics[width=1\linewidth]{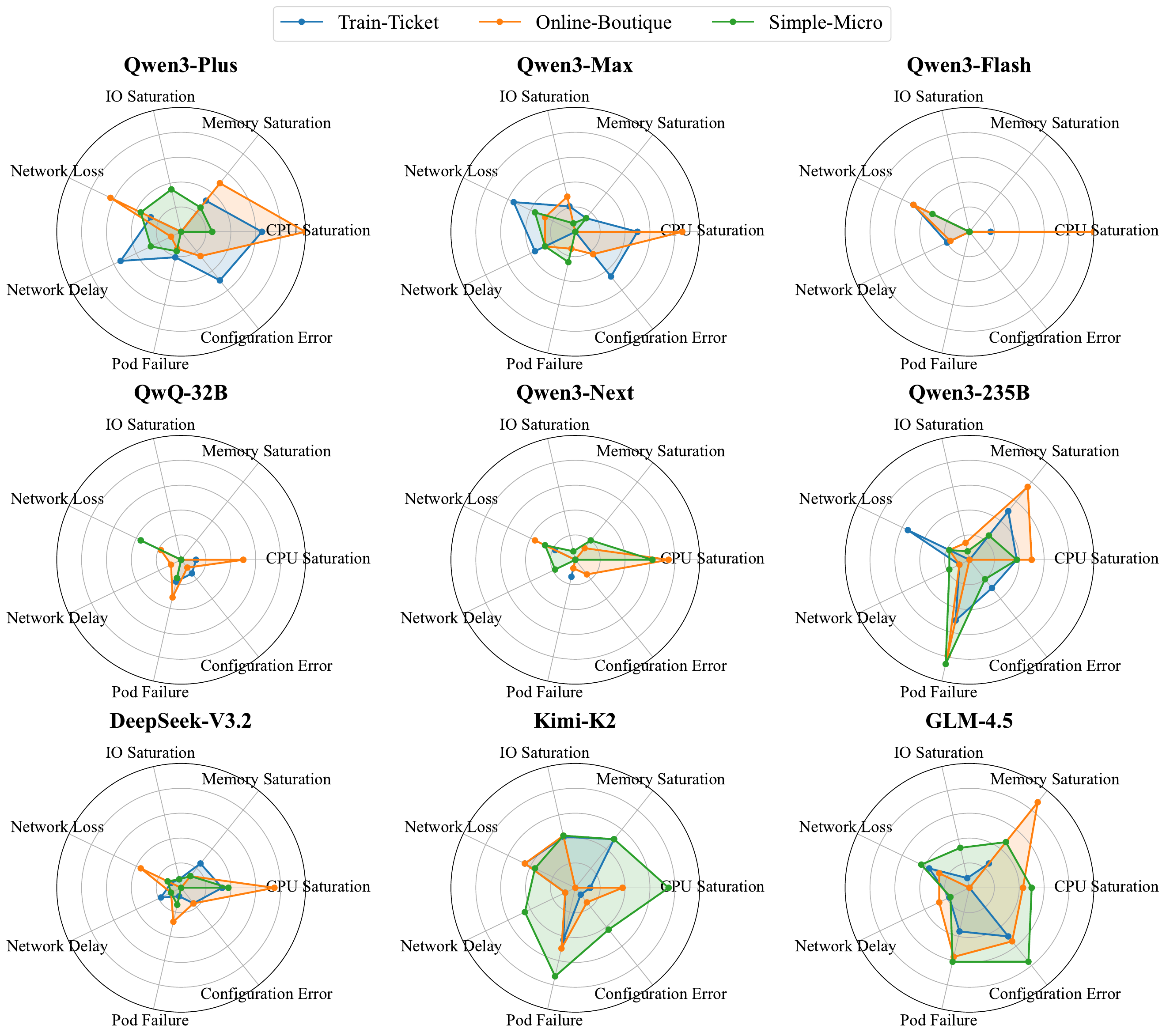}
	\caption{Remediation accuracy of different failure types across nine LLMs}
	\label{fig: class-wise}
\end{figure*}

To further investigate how different LLMs handle heterogeneous fault scenarios, we perform a failure-type-wise evaluation across seven representative fault categories, including CPU Saturation, Memory Saturation, I/O Saturation, Network Loss, Network Delay, Pod Failure, and Configuration Error. As shown in Figure~\ref{fig: class-wise}, the radar charts present the remediation accuracy distribution of nine LLMs across these failure types for the three benchmark systems (Train-Ticket, Online-Boutique, and Simple-Micro).

Overall, the results reveal that model performance varies significantly across failure types. Qwen3-Plus consistently demonstrates the most balanced performance, achieving relatively high accuracy across both system-level and configuration-related failures (e.g., over 60\% on CPU and 50\% on configuration errors in Train-Ticket, and nearly 100\% on CPU faults in Online-Boutique). This indicates that Qwen3-Plus effectively generalizes across different fault sources while maintaining robustness against diverse root-cause patterns. In contrast, smaller or lighter models such as Qwen3-Flash and QwQ-32B show narrow specialization—performing moderately well on network-related issues but struggling severely with resource saturations or configuration anomalies, where accuracies often drop below 20\%.

We also observe that Qwen3-235B and Kimi-K2 display complementary strengths. Qwen3-235B tends to perform better on structured failure types (e.g., memory or pod-related errors), likely due to its large reasoning capacity and long-context understanding, while Kimi-K2 exhibits surprisingly competitive results in simpler microservices such as Simple-Micro, where it achieves above 70\% accuracy on CPU and Pod failures, suggesting efficient reflection under limited contextual complexity.

Another key finding is that network-related faults (Network Loss and Network Delay) remain the most challenging categories across all models. Even top-performing LLMs fail to sustain stable performance in these cases, likely because such issues require reasoning over temporal dependencies and cross-service communication graphs—contexts that are less explicitly represented in current textual traces. Conversely, Configuration Error faults exhibit wide performance variance: while some models (e.g., GLM-4.5 in Simple-Micro) achieve strong accuracy up to 76\%, others almost fail completely, suggesting that LLMs' sensitivity to configuration semantics depends strongly on their pretraining distribution and instruction tuning.

In particular, two fault categories—Pod Failure and Configuration Error—demonstrate distinctive reasoning characteristics. For Pod Failure, most models achieve moderate to high accuracy across microservices, as this type of failure is usually associated with explicit operational symptoms (e.g., missing heartbeat or container restart events) that can be directly captured from logs and metrics. Large models like Qwen3-Plus and Kimi-K2 exhibit near-perfect diagnosis on simple microservices, where causal chains are shallow. However, in complex systems such as Train-Ticket, performance drops sharply due to the propagation of secondary effects—failed pods may cause dependent services to degrade, increasing the difficulty of isolating the original fault.

By contrast, Configuration Error presents a different challenge: rather than manifesting as observable resource anomalies, it often introduces subtle behavioral inconsistencies (e.g., endpoint mismatch or environment variable mis-specification) that require symbolic reasoning and understanding of deployment semantics. Here, reasoning depth and internal reflection play a decisive role. Models equipped with explicit reflection mechanisms (e.g., ThinkRemed) demonstrate more stable performance, as they can iteratively re-evaluate generated hypotheses to eliminate misleading explanations. Nonetheless, even under ThinkRemed, accuracy rarely exceeds 60\%, revealing that configuration-level reasoning remains a fundamental bottleneck for LLM-based remediation systems.

Taken together, these observations emphasize that MicroRemed’s failure-type-wise benchmark successfully exposes heterogeneous reasoning challenges inherent in microservice fault localization. Moreover, it highlights that current LLMs—even strong general-purpose ones—still lack consistent generalization across different failure semantics, underscoring the necessity for task-adaptive reasoning and reflection strategies in future remediation-oriented LLM systems.

\end{document}